\documentclass{article}
\providecommand{\anonbuild}{0}
\PassOptionsToPackage{numbers, compress, sort}{natbib}
\ifnum\anonbuild=1
  \usepackage{neurips_2025}
\else
  \usepackage[preprint]{neurips_2025}
\fi
\usepackage[utf8]{inputenc}
\usepackage[T1]{fontenc}
\usepackage{amsmath,amssymb}
\usepackage{graphicx}
\usepackage{booktabs}
\usepackage{array}
\usepackage{xcolor}
\usepackage{listings}
\usepackage{microtype}
\usepackage{tikz}
\usetikzlibrary{positioning,arrows.meta,fit}
\usepackage{url}
\usepackage[hidelinks]{hyperref}
\newcommand{\toolname}{SpikeBill}
\providecommand{\anonbuild}{0}
\newcommand{\coderepo}{\ifnum\anonbuild=1 \url{https://anonymous.4open.science/r/spikebill-ANON}\else\url{https://github.com/snuhcs/spikebill}\fi}
\newcommand{\eid}[1]{}            %

\newcommand{\x}{$\times$}

\title{SpikeBill: An Open Cost Simulator\texorpdfstring{\\}{ } for Large Spiking Neural Networks\texorpdfstring{\\}{ } on
Multi-Chip Neuromorphic Systems}
\hypersetup{pdftitle={SpikeBill: An Open Cost Simulator for Large Spiking Neural Networks on Multi-Chip Neuromorphic Systems}}
\ifnum\anonbuild=0
\hypersetup{pdfauthor={Hyunseok Oh, Youngki Lee}}
\fi

\ifnum\anonbuild=1
\author{Anonymous Author(s)}
\else
\author{Hyunseok Oh\\
  Seoul National University\\
  \texttt{ohsai@snu.ac.kr}
  \And
  Youngki Lee\\
  Seoul National University\\
  \texttt{youngki.lee@snu.ac.kr}}
\fi

\begin{document}

\maketitle

\begin{abstract}
Spiking neural network (SNN) papers report energy as operation counts times 45\,nm per-operation energies,
which leaves out weight-memory reads, spike routing, step synchronisation and static power. We present \toolname{}, an
open simulator of SNN energy and latency on multi-chip neuromorphic systems; its inputs are the model,
its firing rates and a chip file whose values carry sources and, where uncertain, ranges. Summing
expected event counts over a core-and-chip mapping instead of simulating spikes, it handles
$3.35\times10^{8}$ neurons on 1406 chips in three minutes. On held-out Loihi~1 workloads it matches the event-level
simulator SANA-FE within 0.25\% in energy; we have no silicon measurements, and our Loihi~2 energies rest
on one calibration workload. An estimate from event counts on the
fewest chips is 2\x{} low in energy for a spiking vision transformer whose
weight copies spread it over 16\x{} more chips. Three published spiking transformers would cost at least 10--26\x{}
their reported 45\,nm estimates on Loihi~2 under the most favourable assumptions (30--78\x{} at nominal values),
mostly because each synaptic event reads its weight from on-chip memory. Firing rates from a training framework raise a convolutional SNN's
estimated energy by up to 80\% over chip semantics. We release the simulator, chip files, reference
models and a minimal reporting practice for SNN energy.
\end{abstract}

\section{Introduction}
\label{sec:intro}

Most recent spiking-transformer and spiking-language-model papers estimate energy by counting
operations~\cite{zhou2023spikformer,yao2023sdt,yao2024sdtv2,zhou2023spikingformer,zhu2024spikegpt}. They multiply the
number of synaptic operations by the energy of one addition in a 45\,nm process (0.9\,pJ)~\cite{horowitz2014energy}.
In a survey of 17 such papers from 2023--2026, 13 use this convention (10 exactly, 3 in a close variant), and none
reports a measurement on neuromorphic hardware (Appendix~\ref{app:survey}).

The convention counts arithmetic only, but a chip spends energy on more than arithmetic. A neuromorphic chip is a
mesh of cores. For each spike it reads synaptic weights from memory and sends a message to every core that holds a
target neuron, across the chip and, for large models, between chips. At the end of each time step every core waits
for the slowest core or the busiest link, and the chip draws static power for as long as the step lasts. Benchmarks
and hardware studies have pointed out that operation counts do not predict deployed
cost~\cite{yik2025neurobench,yik2025floorline,dampfhoffer2023are,bhattacharjee2024are,yan2024reconsidering}. No open
tool estimates the cost of an SNN that spans many chips.

\toolname{} is such a tool. It takes an SNN, a chip file and the network's firing rates: the inputs of an
operation-count estimate plus the chip. It places the network on cores and chips and returns the
energy and latency per inference, split into synaptic, neuron, network and static terms. Every chip parameter has a
source, and every uncertain one a low--high range, so each result comes with a range. Like the estimate it replaces,
it excludes host and I/O power.

The design problem is scale: the large models we study need 13 to 1406 chips. Event-level simulators such as
SANA-FE~\cite{boyle2025sanafe} follow each spike, which is accurate but slow; they model a multi-chip system only as
one flat mesh without chip-to-chip links and take the mapping from the user. \toolname{} instead works with expected
counts. For each time step it sums the messages, link loads and per-core work over all pairs of cores, and it takes a
maximum only where the hardware waits, so the \emph{step time} is the time of the slowest core or busiest link plus
the barrier. We call this estimate the \emph{full model} and compare it with a
\emph{first-order estimate} that multiplies event counts by per-event energies and ignores the network, as a
spreadsheet would. We test whether the aggregation loses accuracy against SANA-FE and packet-level network
simulation.

Our contributions are:
\begin{itemize}
\item \textbf{An open chip-system cost simulator that scales} (Section~\ref{sec:model}). It maps a network onto
  cores, balances the load, routes spikes over many chips and counts static power. Expected-count aggregation, a
  model compiled once into sparse matrices, and link loads in constant time per flow let it estimate
  $3.35\times10^8$ neurons on 1406 chips in minutes.
\item \textbf{Validation against finer simulators} (Section~\ref{sec:validation}). On held-out workloads the
  estimate is within 0.25\% of SANA-FE in energy; a Loihi~2 calibration agrees with a second workload but fails on a third.
\item \textbf{A test of when the system model matters} (Section~\ref{sec:multichip}). A first-order estimate with
  the same parameters is within 4\% of the full model for a spiking language model but 2\x{} low in energy for a spiking vision
  transformer that weight copies spread over 16\x{} more chips.
\item \textbf{The size of what the convention leaves out} (Section~\ref{sec:transformers}). The 45\,nm numbers
  correctly count a different quantity, and the papers claim no chip deployment. With the papers' own firing rates,
  three spiking transformers would cost at least 10--26\x{} their reported estimate on our Loihi~2 chip file under
  the most favourable assumptions and 30--78\x{} at nominal values, mostly because each synaptic event reads its weight from on-chip memory, which costs about 26\x{} a 45\,nm addition.
\item \textbf{A reporting practice and reference models} (Section~\ref{sec:usage}). Four items make an SNN energy
  number reproducible; a script recomputes 98 of this paper's numbers.
\end{itemize}

\section{Background and related work}
\label{sec:background}

\textbf{How SNN papers report energy.} SNN energy estimates do not place a model on a neuromorphic chip, route its
spikes or include static power, even when they add memory terms. A typical spiking-transformer paper reports
$E = E_\mathrm{MAC} N_\mathrm{MAC} + E_\mathrm{AC} N_\mathrm{AC}$. Here $N_\mathrm{MAC}$ counts the
multiply-accumulates of the first layer, which reads real-valued inputs, and $N_\mathrm{AC}$ the accumulates implied
by the layer firing rates; $E_\mathrm{MAC}=4.6$\,pJ and $E_\mathrm{AC}=0.9$\,pJ come from a 45\,nm
survey~\cite{horowitz2014energy}. Spikformer~\cite{zhou2023spikformer}, the Spike-driven Transformer
(SDT~V1)~\cite{yao2023sdt}, SDT~V2 (Meta-SpikeFormer)~\cite{yao2024sdtv2} and Spikingformer~\cite{zhou2023spikingformer} use this form.
Other analyses add memory accesses~\cite{lemaire2023analytical}, memory and data movement on generic
architectures~\cite{dampfhoffer2023are,yan2024reconsidering}, systolic and compute-in-memory
accelerators~\cite{bhattacharjee2024are} or output writes~\cite{shen2025step}; none models the cores, links and
static power of a neuromorphic chip system. NeuroBench~\cite{yik2025neurobench} separates hardware-independent
metrics, which do not predict deployed cost, from measurements on hardware; a cost simulator estimates the second
from the first for a chip the author does not have, and does not replace a measurement.

\textbf{Neuromorphic chips.} Three properties of neuromorphic chips set their cost, and the 45\,nm convention
misses all three. Digital chips such as Loihi~\cite{davies2018loihi}, Loihi~2~\cite{intel2021loihi2},
TrueNorth~\cite{merolla2014million} and SpiNNaker~2~\cite{gonzalez2024spinnaker2} are meshes of cores, each holding a
few hundred to a few thousand neurons and their incoming synapses. A spike becomes a message to every core that holds
one of its targets; messages travel over a network-on-chip (a mesh of routers) and between chips over slower links.
All cores wait at a barrier after each time step. Hence (i) each synaptic event reads a weight from memory, (ii) the
slowest core or busiest link sets the step time, and (iii) static power is paid per chip for the whole step. On
measured Loihi~2 workloads static power is 58--88\% of the total~\cite{parpart2023lca,shrestha2024efficient}.

\textbf{SNN chip simulators.} SANA-FE~\cite{boyle2025sanafe} is the closest prior work. It is an open time-step
simulator with a YAML chip description, calibrated against Loihi~1 boards to within 2.7--11.7\% in energy and 3--24\%
in latency. It models a multi-chip system only as one flat mesh, without chip-to-chip links,
takes the mapping from the user and excludes static power. The Loihi~2 runtime model of Timcheck et al.~\cite{timcheck2026runtime} and the Floorline model~\cite{yik2025floorline} bound runtime on
one chip; neither publishes predictions on a workload we could reproduce, so we compare with them only
qualitatively. SpikeSim~\cite{moitra2023spikesim} targets compute-in-memory crossbars.

\textbf{Accelerator cost models and mappers.} \toolname{} applies the idea of accelerator cost models to
event-driven chips. Accelergy~\cite{wu2019accelergy}, Timeloop~\cite{parashar2019timeloop} and
Sparseloop~\cite{wu2022sparseloop} estimate DNN-accelerator energy from per-action costs and a mapping; on a
neuromorphic chip the work per step depends on spike activity, and the step ends at a barrier. Loihi
compilers~\cite{lin2018mapping,rueckauer2022nxtf,lava} place networks under the same core limits as \toolname{}, and
SpiNeMap~\cite{balaji2020spinemap}, SENMap~\cite{nembhani2025senmap} and hypergraph
partitioners~\cite{khorasanian2026beyond} optimise placement on other platforms. NIR~\cite{pedersen2024nir} exchanges
models between frameworks and chips but has no cost model. \toolname{} differs in scope, not in mechanism: it
combines mapping, routing and static power at the scale of $10^8$ neurons and checks its aggregation against finer
simulators.

\section{The simulator}
\label{sec:model}

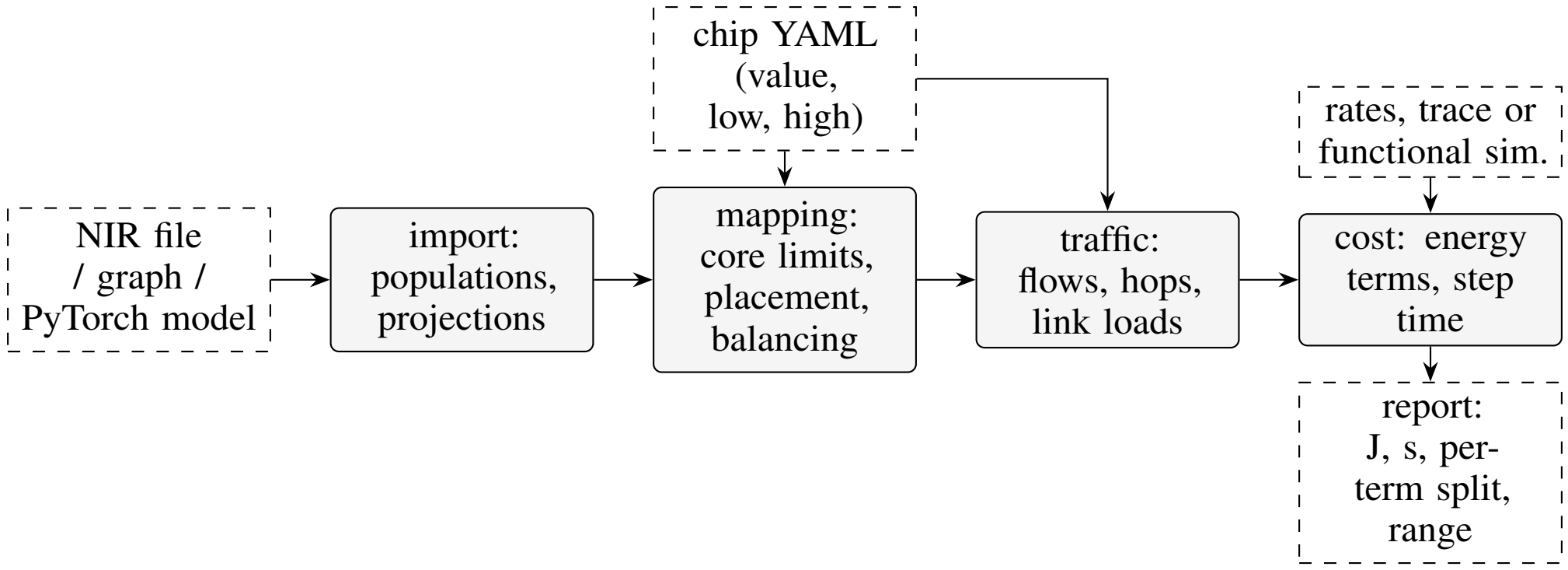
\begin{figure}[t]
\centering
\begin{tikzpicture}[node distance=3mm and 5mm, font=\footnotesize,
  box/.style={draw, rounded corners=2pt, align=center, minimum height=7mm, text width=20mm, fill=gray!8},
  inp/.style={draw, dashed, align=center, minimum height=6mm, text width=20mm},
  ->, >=Stealth]
\node[inp] (nir) {NIR file / graph /\\PyTorch model};
\node[box, right=of nir] (imp) {import:\\populations,\\projections};
\node[box, right=of imp] (map) {mapping:\\core limits,\\placement,\\balancing};
\node[box, right=of map] (tr) {traffic:\\flows, hops,\\link loads};
\node[box, right=of tr] (cost) {cost: energy\\terms, step\\time};
\node[inp, above=of map] (chip) {chip YAML\\(value, low, high)};
\node[inp, above=of cost] (act) {rates, trace or\\functional sim.};
\node[inp, below=of cost] (rep) {report: J, s, per-\\term split, range};
\draw (nir) -- (imp); \draw (imp) -- (map); \draw (map) -- (tr); \draw (tr) -- (cost);
\draw (chip) -- (map); \draw (chip) -| (tr); \draw (act) -- (cost); \draw (cost) -- (rep);
\end{tikzpicture}
\caption{Pipeline of \toolname{}. Dashed boxes are inputs and outputs. The chip file feeds the mapping
(capacities) and the traffic and cost stages (routing, per-event energies and times).}
\label{fig:pipeline}
\end{figure}

\toolname{} turns a model, a chip file and activity into a cost report in four stages
(Fig.~\ref{fig:pipeline}): import, mapping, traffic aggregation and cost evaluation. Appendix~\ref{app:model} gives
the details and the chip parameters; Section~\ref{sec:usage} shows how to run it.

\textbf{Model and chip.} A model is a graph of \emph{populations}, groups of neurons with one neuron model, and
\emph{projections}, the connections between them (dense, sparse, convolutional, one-to-one or explicit edge lists;
convolution and token-parallel layers can share one weight set). It comes from a Neuromorphic Intermediate
Representation (NIR) file~\cite{pedersen2024nir}, a JSON graph, a Python builder or a traced PyTorch network. A chip is
a YAML file that gives the mesh of tiles (each tile is a router with several cores) in $x$, $y$ and, for 3D stacks,
$z$; the capacities of a core; per-event energies and times; barrier times; static power; and the chip-to-chip link.
Every value has a \texttt{source}, and every uncertain value a \texttt{low} and \texttt{high} end; the simulator
reports estimates at the value (the \emph{nominal} estimate) and at both ends (Appendix Tables~\ref{tab:params}, \ref{tab:params-l1}
and~\ref{tab:params-l2}). Our Loihi~2 chip file scales Loihi~1 per-event energies by one factor fitted to a published
measurement (Section~\ref{sec:validation}); we call a system of such chips Loihi-2-class, and time-stepped meshes that store
synapses at the target core Loihi-class.

\textbf{Mapping.} The mapper packs each population into as few cores as the core limits allow, then
rebalances the cores for step time. As on Loihi, a core holds neurons of one population and stores the synapses
that end on them, so its weight memory limits how many neurons it can hold. For each population the mapper chooses
the largest number of neurons per core that fits the neuron, synaptic-memory and input-axon limits (an input axon is
one distinct input source of a core), and fills cores along a snake path (row by row, alternating direction) across tiles and chips. This greedy filling
makes the most active population the bottleneck of every step, because its cores hold the most spiking neurons and
the slowest core sets the step. A balancer therefore sweeps a target time per core, gives busy populations fewer
neurons per core and keeps the mapping with the lowest energy--delay product (EDP). More cores shorten the step but
may add chips and static power. An optional placement search reorders populations on the cores
(Appendix~\ref{app:model}).

\textbf{Traffic.} A spike becomes one message per core that holds one of its targets (unicast, as on Loihi), one
packet copied along a tree of routes (tree multicast), or one packet per destination chip that is copied there (chip
broadcast, documented for Loihi~2~\cite{intel2021loihi2}). For each projection the simulator computes the expected
messages and synaptic events between each pair of cores per spike (the \emph{flows}) from the exact connectivity,
weighted by per-neuron rates when known. It adds each flow to the links on its route, first along $x$, then along
$y$.

\textbf{Energy.} The energy of one time step is the sum of four terms, plus the co-processor's energy when a model
uses one (Appendix~\ref{app:model}):
\begin{align}
E ={}& \underbrace{Y e_\mathrm{syn}}_{\text{synaptic}}
  + \underbrace{N e_\mathrm{upd} + A e_\mathrm{act} + S e_\mathrm{spk}}_{\text{neuron}}
  + \underbrace{M e_\mathrm{out} + H e_\mathrm{hop} + H_\mathrm{c} e_\mathrm{chop}}_{\text{network}} \nonumber\\
  &+ \underbrace{(n_\mathrm{chips} P_\mathrm{chip} + n_\mathrm{cores} P_\mathrm{core})\,t_\mathrm{step}}_{\text{static}}.
\end{align}
The counts are $Y$ synaptic events, $N$ mapped neurons, $A$ active neurons (with input or non-zero state), $S$
spikes generated, $M$ messages sent (one per destination core or packet), and $H$ and $H_\mathrm{c}$ on-chip and
chip-to-chip hops. Each $e_\ast$ is the energy of one such event from the chip file, $P_\mathrm{chip}$ and
$P_\mathrm{core}$ are static power per powered chip and per used core, and $t_\mathrm{step}$ is the step time.
Table~\ref{tab:symbols} maps the symbols to the chip-file names.

\textbf{Step time.} The step time is the slowest of the cores, the network and the optional co-processor (time
$t_\mathrm{cop}$, defined below), plus the barrier:
\begin{equation}
t_\mathrm{step} = \max\big(\max_c t_c,\; t_\mathrm{net},\; t_\mathrm{cop}\big) + t_\mathrm{bar}(n_\mathrm{tiles}, n_\mathrm{chips}).
\end{equation}
A core $c$ has two pipelines that run in parallel, as in SANA-FE's scheduler: one updates neurons and sends their
spikes, the other receives messages and processes synaptic events. Its time is the larger of the two:
\begin{equation}
t_c = \max\big(n_c t_\mathrm{upd} + a_c t_\mathrm{act} + s_c t_\mathrm{spk} + m^\mathrm{out}_c t_\mathrm{out},\;
m^\mathrm{in}_c t_\mathrm{in} + y_c t_\mathrm{syn}\big),
\end{equation}
where $n_c$, $a_c$, $s_c$, $m^\mathrm{out}_c$,
$m^\mathrm{in}_c$ and $y_c$ are the core's mapped neurons, active neurons, spikes, messages sent and received, and
synaptic events, and each such $t_\ast$ is a per-event time. The network time is the time to push the busiest link's
messages through it (its serialisation) plus the latency of the longest route, $t_\mathrm{path}$ (its number of hops times the hop
latency $t_\mathrm{hop}$): $t_\mathrm{net} = \max(L t_\mathrm{link}, L_\mathrm{c}
t_\mathrm{clink}) + t_\mathrm{path}$, where $L$ and $L_\mathrm{c}$ are the message loads of the busiest on-chip and
chip-to-chip link and $t_\mathrm{link}$ and $t_\mathrm{clink}$ their times per message. The barrier time
$t_\mathrm{bar}$ is one value on a single chip (on Loihi~1, a function of the used tiles) and a larger fixed value for
any multi-chip system (Table~\ref{tab:params}). The outer maximum assumes that
compute and communication overlap. Against finer simulators the step time is between 59\% low (4-flit router
queues) and 45\% high (light load); with ample buffers the median error is within $\pm$5\%
(Appendix~\ref{app:val-signed}). Evaluated at mean rates, it understates the mean latency per input when the
slowest core varies by input (by 28\% for SDT~V1; Section~\ref{sec:limits}).

\textbf{Co-processor.} Spiking attention multiplies two spike tensors (such as $QK^\top$), which a neuron core does
not do. \toolname{} costs such products, and real-valued element-wise operations, either on a co-processor with time
$t_\mathrm{cop}$ or as AND-gate neurons, which fire when both inputs spike in the same step. Neither Loihi chip has
such a co-processor; its parameters are our assumptions (Appendix~\ref{app:params}).

\textbf{Execution and static power.} A \emph{time step} is a step of the SNN; a \emph{chip step} is one step of the
hardware, which ends at the barrier. In pipelined execution, as on Loihi, each chip step moves spikes one layer forward,
so an inference of $T$ time steps takes $T+D$ chip steps for a network of depth $D$. Lockstep execution ignores this
layer delay and counts $T$ chip steps. Static power counts every powered chip by default. Appendix~\ref{app:model}
lists the other accounting choices; two published reports of one Loihi measurement differ 7\x{} because one
counts the activity of the chip's embedded x86 processors and the other does not.

\textbf{Activity.} The cost model needs the spike counts of each population: rates a paper reports, a recorded
trace, or the output of \toolname{}'s GPU functional simulator. The simulator runs the graph in \emph{chip}
semantics, as on Loihi (a spike reaches its targets one step later, the membrane is quantised, reset is hard), or in
\emph{framework} semantics, as in training frameworks (a spike reaches its targets in the same step, reset is
subtractive). The two semantics give different rates in later layers (Section~\ref{sec:semantics}). Given the
activity, the estimate is deterministic; variation across inputs enters only through the activity.

\textbf{How the estimate scales.} Four techniques let \toolname{} handle $3.35\times10^8$ neurons on 1406
chips in minutes, where an event-level simulator follows every spike on one chip. The two speed-ups measured below
are exact.

\emph{(a) Expected counts, maxima only where the hardware waits.} \toolname{} does not follow individual spikes. It
sums expected messages, synaptic events and link loads over core-to-core flows and takes a maximum only at the
slowest core, the busiest link and the barrier, where the hardware waits.

\emph{(b) Compile once, evaluate as one matrix product.} Every quantity of Eqs.~(1)--(3) except the maxima is linear
in the per-population spike counts. \toolname{} therefore compiles graph, chip and mapping once into sparse
matrices; $T$ steps or $N$ inputs are then one sparse matrix product followed by the maxima. On SDT~V1 (164 chips) a step
takes 0.81\,ms instead of 0.90\,s, with identical results; compiling takes 0.11\,s once. This makes per-input spreads
over thousands of inputs, and re-evaluation inside a training loop, affordable.

\emph{(c) Link loads in constant time per flow.} Adding a flow's load to every link on its route would cost its path
length, which grows with the mesh. \toolname{} instead adds the load at the start of the route's straight legs and
subtracts it after their ends in one difference array per direction, separately for on-chip, chip-to-chip and
vertical links; one cumulative sum then gives every link's load. On a 1406-chip mesh (routes of 151 hops on average) this is 31\x{} faster than walking each
route for $10^6$ flows, with identical loads. The three multicast modes change only the flows.

\emph{(d) Parallel balancing, placement checked with the cost model.} The balancer evaluates its 16 target step
times in parallel workers, and every candidate placement (graph order, spectral, Kernighan--Lin, annealed;
Appendix~\ref{app:model}) is scored with the cost model, not a proxy.

A full estimate of the largest model, mapping and balancing included, takes 179\,s on eight workers; for the cost
given activity, \toolname{} is 57--3268\x{} faster than SANA-FE (Appendix~\ref{app:val-speed}).

\textbf{Two simpler baselines.} We call \toolname{}'s estimate the \emph{full model} and compare it with two simpler
estimates that use the same chip file. The \emph{first-order estimate} multiplies event counts by per-event energies,
charges the static power of the fewest chips that hold the neurons and weights, spreads the work evenly over their
cores, and has no network term. The \emph{roofline} keeps the full mapping and each core's time but spreads traffic
evenly over the links: it sees the slowest core, not the busiest link.

\section{Validation}
\label{sec:validation}

We validate the aggregation against finer simulators, because we have no silicon measurements of our own. \toolname{}
matches an event-level simulator on held-out Loihi~1 workloads and a packet-level simulator on multi-chip traffic; a
Loihi~2 calibration agrees with a second published workload but not with a third. Table~\ref{tab:validation} summarises the
tests. Each had its pass condition committed before the run, except where marked post-hoc; Appendix~\ref{app:prereg}
lists every test and Appendix~\ref{app:validation} gives the details.

\begin{table}[t]
\centering
\caption{Validation summary. Errors are (\toolname{} $-$ reference) / reference. BookSim2 is a cycle-level network
simulator; a virtual channel (VC) is a separate router queue. PilotNet (a driving CNN) and N-DNS (an audio denoiser)
are published Loihi~2 measurements. Pass conditions were fixed
before the run except where marked. Parentheses: Wilson 95\% interval (Appendix~\ref{app:val-stats}).
$^\dagger$Passes with two virtual channels, chosen after the first runs (post-hoc); fails with the pre-registered one.
$^\ddagger$A seed-0 prototype was seen before registration; the criterion is unchanged and the scored runs used new seeds or new graphs.}
\label{tab:validation}
\footnotesize
\setlength{\tabcolsep}{3pt}
\begin{tabular}{@{}p{28mm}p{31mm}p{28mm}p{35mm}l@{}}
\toprule
Test & Workloads & Pass if & Result & Verdict \\
\midrule
Model vs SANA-FE & 8 held-out configurations & energy $\pm$10\%, latency $\pm$20\% & energy $\le$0.25\%; latency $\le$6.2\% & pass \\
Functional sim.\ vs SANA-FE & DVS net, 5 mappings & same spikes; energy $\pm$10\%, latency $\pm$20\% & same spikes; latency $\le$7.5\% & pass \\
Network vs packet sim. & DVS, synthetic; 1--16 chips & $\le$25\% of steps off by $>$20\% & 0\% (DVS), 17\% (synthetic) & pass \\
Chip broadcast vs packet sim. & 22 steps (11 configs, 2 seeds), 1--57 chips & same & 0\% & pass$^\ddagger$ \\
Network, 4-flit queues & synthetic & $\ge$75\% of steps within 20\% & 57\% (16 flits: 78\%, post-hoc) & fail \\
Network vs BookSim2 & $k\times k$ mesh, $k=8$--32, 4 patterns, 3 loads & $\ge$75\% within 20\% & 2 VCs: 30/36 (68--92\%); 1 VC: 15/36 & mixed$^\dagger$ \\
Loihi 2 scale, two workloads & PilotNet and LCA fitted separately & ratio $\le$1.5 & 1.15 & pass \\
Loihi 2 transfer & calibrated on PilotNet, predicts N-DNS & within 2\x{} & 11--22\x{} low & fail \\
\bottomrule
\end{tabular}
\end{table}

\textbf{Event-level simulation (Loihi~1).} On eight held-out configurations, \toolname{} is within 0.25\% of
SANA-FE~\cite{boyle2025sanafe} in energy and within 6.2\% in latency. A configuration is one network with one mapping;
the networks are SANA-FE's public dynamic vision sensor (DVS) gesture network and random networks. Both simulators
used the same chip parameters, mappings and activity. We developed the model on three DVS mappings and nine random
networks. The latency figure comes after a post-hoc fix of the busiest-link term, which had used the per-hop
latency as the time per message (no criterion changed; Appendix~\ref{app:prereg}). With shared activity, the 0.25\% energy agreement mainly shows that we count events as SANA-FE does.

Our functional simulator, run in chip semantics on five mappings of the DVS network, produces the same 365,277 spikes
as SANA-FE; with this activity the energy is unchanged and the latency is 1.9--7.5\% lower. SANA-FE itself is within 2.7--11.7\% in energy and 3--24\% in latency of Loihi~1 silicon, so these
tests bound the error of our aggregation, not our error against silicon.

\textbf{Packet-level simulation.} The network term matches a packet-level simulation except at light load, where it
over-predicts. SANA-FE has no chip-to-chip links, so for multi-chip traffic we compare with our own event-driven packet-level
simulator, which follows every packet through first-in-first-out link queues and ships with the release. On DVS
traffic, none of 50 sampled steps is outside $\pm$20\%. On synthetic traffic over 4 and 16 chips, 9 of 54 steps are
outside, all at light load (4 packets per core): the busiest link is then almost idle, and adding a full route
latency over-predicts them by 21--31\%. At heavy load (64 packets per core) the error is at most 4\%. With chip broadcast,
none of 22 steps is outside $\pm$20\%. On SpikeGPT, mapped greedily on 13 chips, six sampled steps are within 8\%
(criterion set post-hoc).

\textbf{Finite buffers and a cycle-level router.} The network term assumes infinite router buffers; two further tests show where it stops holding. The pre-registered test used 4-flit queues (a flit is the unit a router forwards;
here one packet) and failed: 57\% of synthetic-traffic steps were within $\pm$20\%, against the 75\% required. With
16-flit queues 78\% are (post-hoc), so the term should not be used for queues shorter than 16 flits. Against the cycle-level network simulator
BookSim2~\cite{jiang2013booksim}, 30 of 36 configurations are within $\pm$20\%, but only with two virtual channels per
router, chosen after the first runs; with the pre-registered one, 15 of 36 are
(Appendix~\ref{app:val-packet}). Loihi's router is unpublished.

\textbf{Loihi~2: two workloads agree, a third does not.} Loihi~2 per-event energies are not public, so we scale
the Loihi~1 energies by one factor $s$, fitted to the on-chip counters of a published PilotNet run (a driving
CNN)~\cite{lavadl2023pilotnet}; the chip file uses $s=0.65$ (range 0.33--1.0, Appendix~\ref{app:val-loihi2}). A
pre-registered fit to a one-layer LCA network~\cite{parpart2023lca}, with activity read from that paper's text, gives
$s=0.56$, within 15\% of PilotNet's 0.64. Intel's N-DNS audio denoiser~\cite{shrestha2024efficient} does not fit: its
predicted dynamic power is 11--22\x{} too low (0.16--0.31 against 3.48\,mW), and its published static/dynamic split
cannot be a whole-chip split. Omitted I/O and processor power may also contribute (untested).

\section{Case study: when the full model changes the answer}
\label{sec:multichip}

The full model changes the estimate when weight memory spreads a model over many more chips than its work
needs; it does not when the cost is dense work on a few chips. We show one case of each, both on Loihi~2 at nominal parameters with balanced mapping, pipelined
execution and tree multicast, and compare each with the first-order estimate of Section~\ref{sec:model}.

\textbf{SpikeGPT: a few chips, dense work.} For SpikeGPT the first-order estimate is within 4\% of the full model.
SpikeGPT~\cite{zhu2024spikegpt} is a 216M-parameter spiking language model with a public checkpoint; we measured its
activity on WikiText-103 with a port of the released code (Appendix~\ref{app:limits}). Its linear layers read the
real-valued residual stream, not spikes: only $3.2\times10^{-5}$ of its synaptic events are driven by spikes. We cost
each real value as a graded spike sent every step, which Loihi~2 supports~\cite{intel2021loihi2}, at one synaptic
event per outgoing weight. Its cost is therefore dense work, which the first-order estimate counts. The full model maps it on 16 chips at 3.59\,mJ
per token (1.84--11.4\,mJ over the parameter range); the first-order estimate gives 3.45\,mJ on 11 chips. Synaptic
events are 88\% of the energy. The step is set by the assumed co-processor (Section~\ref{sec:model}), which runs the real-valued element-wise operations.

\textbf{SDT~V2-55M: hundreds of chips, network-bound.} For SDT~V2-55M the first-order estimate
is 2\x{} low in energy. SDT~V2 is published in a 31M and a 55M
size~\cite{yao2024sdtv2}; SDT~V2-55M widens every layer of SDT~V2-31M by 4/3. Our graph matches the released 55M
checkpoint layer for layer (Appendix~\ref{app:limits}). We assume the published firing rates of SDT~V2-31M and test this below.

The full model needs 506 chips, because a convolution's shared weights are copied into every group of cores that
holds part of the layer. It costs 2.51\,J per image (1.34--13.6\,J) and 10.7\,ms. The network sets the step: the busiest
link needs 94.1\,\textmu s per step, the busiest core 20.0\,\textmu s. Static power over 506 chips is half of the energy.
The first-order estimate, which has no network term and needs only 32 chips, gives 1.24\,J. The 2\x{} gap does not depend
on the assumed rates: with rates measured on ImageNet with the released checkpoint, the full model needs 435 chips and
costs 2.33\,J, 1.93\x{} the first-order estimate at the same rates. Appendix~\ref{app:limits} gives the effect of
other rates and mappings.

\textbf{What sets the first-order error.} The first-order error grows with the number of chips: 4\% for SpikeGPT on 16 chips, 1.3--1.5\x{} for the three transformers of Section~\ref{sec:transformers}
on 164--189 chips, and 2\x{} for SDT~V2-55M on 506 chips. Most of the gap is static power on the chips that weight copies add: about 80\% for the
transformers (164--189 chips against 17--24) and 92\% for SDT~V2-55M (506 against 32). The network term is
3\% of its energy but sets its step time. The roofline of Section~\ref{sec:model} misses the busiest link (3.2\x{} low in latency for SDT~V2-55M at the assumed rates; Appendix~\ref{app:limits}).

\section{Case study: what the 45\,nm estimate leaves out}
\label{sec:transformers}

The 45\,nm estimates of three published spiking transformers count arithmetic, not chip energy, and the papers claim
no chip deployment. On a Loihi-2-class system, with the papers' firing rates and nominal parameters, the models would
cost 30--78\x{} their reported estimate per inference, mostly because each synaptic event reads its weight from on-chip memory, which costs about 26\x{} a 45\,nm addition; under a floor
that no real mapping reaches (every weight stored once, every parameter favourable), 10\x{} or more.

\textbf{Setup.} We hold each model's behaviour fixed and change only the cost accounting. The models are SDT~V1
8-512 (8 blocks, width 512)~\cite{yao2023sdt}, Spikingformer 8-512~\cite{zhou2023spikingformer} and SDT~V2-31M (31.3M
parameters)~\cite{yao2024sdtv2}, all at $T=4$ time steps on ImageNet, with the per-layer firing rates each paper
publishes. (With AND-gate attention, Spikingformer needs 1406 chips; Appendix~\ref{app:transformers}.) The chip is Loihi~2 with 8-bit weights, tree multicast and balanced mapping; products of spike tensors in
attention go to the assumed co-processor. Static energy is charged over the full latency of one inference ($T+D$
chip steps) with every powered chip counted, which is conservative for back-to-back inferences. With the papers'
convention our graphs reproduce two of the three reported estimates (Appendix~\ref{app:transformers}).

\begin{table}[t]
\centering
\caption{From the papers' 45\,nm estimate to the chip estimate (mJ per image, ImageNet, $T=4$). Each row adds one correction
to the previous one; row (c) counts one event per input spike and target synapse, as a chip does; row (c$'$) is a
45\,nm reference, not a step of the chain. Loihi~2 nominal parameters, balanced
mapping, co-processor attention. Range: all chip parameters at their low or high end together (the high end includes a single co-processor
unit), mapping re-balanced at each end; joint: 5th--95th percentile of a Monte Carlo over all parameters.}
\label{tab:transformers}
\small
\begin{tabular}{@{}lrrr@{}}
\toprule
Step & SDT V1 & Spikingformer & SDT V2-31M \\
\midrule
(a) reported in the paper & 4.5 & 7.5 & 32.8 \\
(b) paper's formula, reproduced & 4.2 & 7.5 & 22.4 \\
(c) events as a chip counts them $\times$ 0.9\,pJ & 8.5 & 7.4 & 24.2 \\
(c$'$) (c) + 45\,nm SRAM read, 32\,KB & 169 & 145 & 533 \\
(d) events $\times$ chip synaptic energy & 185 & 158 & 588 \\
(e) + neuron updates and network & 249 & 221 & 721 \\
(f) + static power (= chip estimate) & 353 & 317 & 970 \\
\midrule
Chip estimate / reported & 78\x{} & 42\x{} & 30\x{} \\
Range over chip parameters & 38--266\x{} & 20--202\x{} & 15--97\x{} \\
Joint uncertainty, 5th--95th percentile & 69--180\x{} & 39--121\x{} & 26--68\x{} \\
Chips & 164 & 168 & 189 \\
\bottomrule
\end{tabular}
\end{table}

\begin{figure}[t]
\begin{minipage}[c]{0.5\linewidth}
\includegraphics[width=\linewidth]{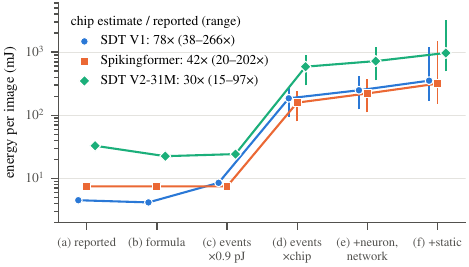}
\end{minipage}\hfill
\begin{minipage}[c]{0.47\linewidth}
\caption{From the papers' 45\,nm estimate to the chip estimate for the three spiking transformers (steps (a)--(f) of
Table~\ref{tab:transformers}, log scale). The largest step is (c)$\to$(d), the energy of a synaptic event. Whiskers:
every chip parameter at its low or high end, mapping re-balanced.}
\label{fig:waterfall}
\end{minipage}
\end{figure}

\textbf{Four corrections} lead from a reported estimate to the chip estimate (Table~\ref{tab:transformers},
Fig.~\ref{fig:waterfall}), and the second is by far the largest. \emph{Counting events} (b$\to$c) changes the number
by up to 2.0\x{}: SDT~V1 weights the Q/K/V projections by their output rates; a chip processes one event per input spike.

\emph{Energy per synaptic event} (c$\to$d) multiplies the number by 21--24\x{}. A chip event reads a weight from
synaptic memory, which costs about 23\,pJ in our calibration, against 0.9\,pJ for a 45\,nm addition. If every event reads its own weight word, as event-driven
cores do, one 64-bit read from a 32\,KB on-chip memory (SRAM) at 45\,nm per event of row (c) gives row (c$'$), 91\% of row (d).

\emph{Neuron updates and the network} (d$\to$e) add 1.2--1.4\x{}, and \emph{static power} of 164--189 chips over
the inference latency (e$\to$f) another 1.3--1.4\x{}. In the chip estimate, synaptic events take 50--61\% of the
energy and static power 26--30\%. The corrections also reverse a ranking: by the reported estimates SDT~V1 is cheaper
than Spikingformer (4.5 vs 7.5\,mJ); on the chip it costs more (353 vs 317\,mJ).

\textbf{What the result depends on.} The ratios stay above 10\x{} under every alternative we tried: attention
built from AND-gate neurons, greedy mapping, joint sampling of all chip parameters (26--180\x{}), and a floor that no
real mapping reaches (Appendix~\ref{app:transformers}). Two small trained SNNs on one chip show the same gap,
78--130\x{} (descriptive).

\textbf{What needs the full model.} The energy gap does not need the full model: the first-order estimate is
already 22--55\x{} the reported estimates, and the full model adds 1.33--1.45\x{}, mostly static power on the 164--189
chips that weight copies need instead of 17--24 (Section~\ref{sec:multichip}). The gap rests on the per-event
synaptic energy, as noted before for generic architectures~\cite{lemaire2023analytical,dampfhoffer2023are,yan2024reconsidering}.
Chip count and latency need the full model: on few chips the first-order latency is 1.8--2.0\x{} the mapped one.

\textbf{What the result does not show.} Nothing here bears on the papers' comparisons with ANNs, which we did not
estimate.

\section{Case study: where activity comes from}
\label{sec:semantics}

Firing rates from a training framework can inflate a chip estimate by tens of percent. Published rates usually
come from such a framework (Section~\ref{sec:model}). In a framework, a
spike reaches the next layer in the same step, and subtractive reset keeps the membrane's excess after a spike, so
neurons in later layers fire more often. \toolname{}'s functional simulator runs the same weights and inputs in framework
and chip semantics (pre-registered criterion: more than 10\% in energy or 20\% in latency).

The effect is large when most synaptic events occur after the first layer. Framework semantics raise the energy by
74--80\% on a DVS convolutional network, by 31\% on Loihi~1 and 13\% on Loihi~2 for a trained three-layer
convolutional SNN, and by under 1\% for a trained MLP (Appendix Table~\ref{tab:semantics}). In the convolutional
networks the rates of deeper layers rise by 1.3--3.5\x{}; only the DVS network's output layer falls, to 0.49\x{}. In the MLP most events come from the input layer, whose
spikes are the same in both. SDT~V1, which uses hard reset, changes by only 2\% (Appendix~\ref{app:transformers}).

Accuracy does not reveal the effect (94.1\% vs 92.2\% on MNIST for the three-layer SNN), so networks with
subtractive reset should be simulated in chip semantics, as the functional simulator does.

\section{Using the simulator and reporting the result}
\label{sec:usage}

\textbf{One call.} A model, a chip and the firing rates give a report in one call:
\begin{lstlisting}
from spikebill import Simulator
sim = Simulator("model.nir", chip="loihi2", multicast="tree")
rep = sim.run(rates={"conv1": 0.05}, steps=4, balance=True, ranges=True)
print(rep.summary())      # J, s, energy split, term that sets the step
print(rep.paper_block())  # citable block for the reporting practice
\end{lstlisting}
Appendix~\ref{app:datasheet} describes the command line, a PyTorch tracer (\texttt{Simulator.from\_torch}), per-input
spreads and per-population attribution. A new chip is one YAML file.

\textbf{A minimal reporting practice.} An SNN paper can replace the operation-count estimate with a chip-level one
by stating four things: (1) the chip file and simulator version; (2) the value and the low--high range of the result;
(3) the mapping options (balanced or greedy, multicast or unicast, weight precision, static accounting); and (4) the
source of the activity (framework or chip semantics, measured on which data). \texttt{rep.paper\_block()} prints most of these items as a citable block (Appendix~\ref{app:datasheet}). To compare across papers, we
recommend the settings of the reference estimates shipped with the tool (\texttt{zoo.REFERENCE\_SETTINGS}) and
stating any departure.

\textbf{Availability and reproduction.} \toolname{} is released under the Apache-2.0 license at
\coderepo{}, with chip files, documentation, the packet-level simulator and tests. The
models of this paper ship as a model zoo with their rates and reference estimates
(\texttt{Simulator("zoo:sdt\_v1\_8\_512")}), and one script recomputes 98 printed numbers. Appendix~\ref{app:datasheet}
states the intended use, maintenance and versioning.

\section{Limitations}
\label{sec:limits}

\textbf{No silicon validation of our own.} The model agrees with SANA-FE, which is itself within 2.7--11.7\% in energy and
3--24\% in latency of Loihi~1 silicon~\cite{boyle2025sanafe}, and with our packet-level simulator. It passes the BookSim2 test only with a router setting chosen post-hoc. The Loihi~2 per-event energies
rest on one scale factor fitted to one published workload; a second workload gives a factor within 15\%, a third
does not fit, and for I/O-heavy tasks our Loihi~2 energies are lower bounds.

\textbf{Proxy, assumed and omitted terms.} The chip-to-chip hop energy is assumed, and the multi-chip
barrier, a Loihi~1 value, adds to every step of a multi-chip model. The model has no term for I/O, embedded processors or
the host. We attribute the N-DNS miss to it, untested; in a published 24-chip model this term is about 0.5\,W
per chip (Appendix~\ref{app:abreu}).

\textbf{Mean rates.} Because the step is a maximum over cores, the cost at mean firing rates is at most the mean
cost per input. For SDT~V1 on 5,000 ImageNet images the gap is 10\% in energy and 28\% in latency. For SDT~V2-55M and the wider Spikingformer 8-768 it is under 0.1\%.

\textbf{Mapping.} The mapper is not optimal. Within a core, a
convolution's weights are stored once for all positions it holds (Loihi~2's shared synapses); cores holding other
positions store copies, because synaptic memory is per core (the copy-free case is the floor of
Appendix~\ref{app:transformers}).
A placement search lowers SDT~V2-55M's EDP by up to
38\%, so its numbers are upper estimates (1.15\x{} in energy, 1.41\x{} in latency). Splitting populations over core blocks
shortens SDT~V2-31M's step by 13\% (Appendix~\ref{app:limits}).

\textbf{Evaluative role and scope.} \toolname{} supports comparisons on a stated Loihi-class chip file (of models, mappings,
parameters and cost terms, and of operation counts against chip estimates), not predictions of measured power. The large models are five spiking vision transformers (three families) and one spiking
language model; large recurrent, audio and event-camera models are untested.

\section{Conclusion}
\label{sec:conclusion}

\toolname{} estimates the chip-system cost of SNNs with up to $3.35\times10^8$ neurons in minutes; chip count, link
hotspots and static power set much of the cost of large models. Three published spiking transformers would cost at
least 10--26\x{} their reported estimate on our Loihi~2 chip file, and 30--78\x{} at nominal values, mostly because each
synaptic event reads its weight from on-chip memory.

\ifnum\anonbuild=0
\paragraph{Acknowledgments.} We used an AI coding assistant (Claude Code, Anthropic) for software implementation,
experiment scripts and drafting text. The authors directed the research, verified all results and take full
responsibility for the content.
\fi

\bibliographystyle{plainnat}
\bibliography{refs}

\appendix
\section{Simulator details}
\label{app:model}

\textbf{Import.} The NIR front end converts Affine, Conv, SumPool, Flatten and leaky integrate-and-fire nodes into
populations and projections, composing chains such as Conv$\to$SumPool$\to$Flatten$\to$Affine into exact edge lists.
On a published SpiNNaker~2 DVS-gesture network it recovers the reported 7,859 neurons and 909,952 synapses.

\textbf{Chip parameters.} Table~\ref{tab:params} lists the main parameters of the two chips we use; the full chip
files are in Tables~\ref{tab:params-l1} and~\ref{tab:params-l2}. Loihi~1 values come from SANA-FE's silicon
calibration~\cite{boyle2025sanafe}, with Davies et al.~\cite{davies2018loihi} giving the range. Loihi~2 per-event
energies are not public; we scale the Loihi~1 values by 0.65 (range 0.33--1.0), the factor that fits one Loihi~2
measurement (Appendix~\ref{app:val-loihi2}). The chip-to-chip hop energy is not public for any chip; we assume ten
times Loihi~1's on-chip hop (3.6\,pJ, so 36\,pJ per chip-to-chip hop) for both chips, with a range of one to one hundred times.

\begin{table}[h]
\centering
\caption{Main chip parameters (value; low--high where uncertain; Loihi~2 chip file v2026.11.5-e6, whose values equal
those of v2026.11.2-e6 used for the runs; the suffix e6 marks energies scaled by the fitted factor). A hop is one
router-to-router transfer of a message; link times are per message. [C] SANA-FE
silicon calibration; [P] published (Loihi~2: technology brief~\cite{intel2021loihi2}; multi-chip
barrier:~\cite{frady2020pohoiki}); [D] derived from published totals (Loihi~1 static: 768-chip system
power~\cite{frady2020pohoiki}; Loihi~1 chip link: Loihi~2 brief transfer rate); Loihi~2 static is [P], from board
measurements~\cite{lavadl2023pilotnet,abreu2025neuromorphic,parpart2023lca}; [S] scaled from Loihi~1 (Appendix~\ref{app:val-loihi2}); [L1] Loihi~1 value as proxy; [A] assumed.}
\label{tab:params}
\small
\begin{tabular}{@{}lll@{}}
\toprule
Parameter & Loihi 1 & Loihi 2 \\
\midrule
Tiles ($x\times y$) $\times$ cores per tile & $8\times4\times4$ & $8\times4\times4$ [A] \\
Neurons per core (max) & 1024 [P] & 8192 [P] \\
Synaptic memory per core & 128\,KB [P] & 128\,KB$^\ast$ [P] \\
Synaptic event energy (pJ) & 35.5 (23.6--35.5) [C] & 23.1 (11.7--35.5) [S] \\
Neuron update, every neuron (pJ) & 51.2 [C] & 33.3 (16.9--51.2) [S] \\
Spike message out (pJ) & 111 (1.7--111) [C] & 72.2 (36.6--111) [S] \\
On-chip hop (pJ) & 3.6 (3.0--4.2) [C] & 2.34 (1.19--3.60) [S] \\
Chip-to-chip hop (pJ) & 36 (3.6--360) [A] & 36 (3.6--360) [A] \\
Synaptic event time (ns) & 3.8 [C] & 0.76 (0.76--3.8) [P] \\
Message in, per core (ns) & 16 [C] & 16 (3.2--16) [A] \\
Tile link, per message (ns) & 0.29 [P] & 0.29 (0.29--5.3) [L1] \\
Chip link, per message (ns) & 5.4 [D] & 1.25 (1.25--5.0) [P] \\
Barrier, one chip / multi-chip (\textmu s) & 1.8 [C] / 14.3 [P] & 1.8 (0.2--1.8) / 14.3 (1.8--16.2) [L1] \\
Static power per chip (W) & 0.07 (0--0.07) [D] & 0.23 (0.15--0.55) [P] \\
\bottomrule
\end{tabular}
\\[1mm]{\scriptsize $^\ast$ upper bound of the synapse partition of 192\,KB soft-partitioned memory~\cite{intel2021loihi2}.}
\end{table}

\textbf{Mapping.} The mapper reports violations of the core limits that no choice of neurons per core can fix; it
does not drop them. Chips form a near-square board, or a stack of dies for 3D packages, and a user can pass an
explicit placement instead. For each target time per core, the balancer multiplies each population's neurons per core by
(target time / its busiest-core time) and re-maps (16 targets, two iterations); it keeps the target that minimises EDP, energy or latency.

\textbf{Placement search.} \texttt{place=True} changes the order in which populations fill the cores, so that
populations with heavy traffic between them sit close together. It tries four kinds of order: the graph order; a spectral
order of the graph of traffic between populations (sorted by its Fiedler vector); orders from recursive
Kernighan--Lin bisection, which splits that graph into halves with little traffic between them; and simulated
annealing of the order on a fast proxy of energy and busiest-link time, calibrated against the full model. Every candidate is checked with the full model and the best is kept. \texttt{balance(place=True)}
searches placement inside the balancer for every candidate partition; \texttt{run(place="split")} also allows a
population to occupy non-contiguous blocks of cores. Appendix~\ref{app:limits} reports what the search changes.

\textbf{Traffic.} Optional relay populations forward the spikes of a high-fan-out population, which reduces the
copies the source core must send. Flows come from the exact connectivity, because a closed-form estimate mis-estimates
messages by up to 6\x{} (0.16--3.9\x{} the exact count).
Link loads accumulate with difference arrays, so each flow costs $O(1)$ regardless of its path length. Vertical hops
$H_\mathrm{v}$ and vertical links (energy $e_\mathrm{vhop}$, time $t_\mathrm{vlink}$) enter the energy and the network
time as the chip-to-chip terms do. Spikes are spread over a population's cores in proportion to their neuron counts
or measured activity.

\textbf{Neuron updates.} $N$ counts every mapped neuron, because Loihi's dendrite unit (the circuit that updates neuron state) updates all neurons of a core
every step~\cite{davies2018loihi}. Loihi~2's behaviour is not documented, so a chip option restricts
updates to neurons with input or state; it lowers the transformer estimates by 1.5--2.2\%.

\textbf{Static accounting.} Static power can count whole powered chips (the default, a system view), only the used
fraction of cores (common in reports from Intel's Loihi software and from TrueNorth), or nothing (SANA-FE's dynamic-only view). Such choices
change published numbers by large factors: two reports of the same Loihi keyword-spotting measurement,
270\,\textmu J~\cite{blouw2019benchmarking} and 37\,\textmu J~\cite{yan2021comparing} per inference, differ 7\x{}
because the second removes about 70\,mW drawn by the chip's embedded x86 processors, which manage the mesh
interface. Every number in this paper states its accounting.

\textbf{Core timing.} The two sides of a core overlap in the model because summing them over-predicted latency by up
to 67\% in our first validation and by up to 86\% in the held-out ablation (Appendix~\ref{app:val-sanafe}); the chip file can select either.

\textbf{Co-processor.} With $O$ operations and $M_\mathrm{cop}$ messages to the co-processor in a step, the model
adds $E_\mathrm{cop} = O\,e_\mathrm{op} + M_\mathrm{cop}\,e_\mathrm{out}$ to the energy and
$t_\mathrm{cop} = O\,t_\mathrm{op}$ to the step maximum, with assumed $e_\mathrm{op}=0.03$\,pJ and $t_\mathrm{op}=0.125$\,ns
(Appendix~\ref{app:params}). $O$ and $M_\mathrm{cop}$ scale with the spikes of the
populations that feed the product, as measured when the model is traced. Results return as spikes of the product's
output population, which the network term costs like any other message. The AND-gate alternative maps the product
as neurons that fire when both inputs spike, which the equations of Section~\ref{sec:model} cover directly.

\textbf{Implementation.} A test checks that the compiled matrix path equals the direct per-step code.
\texttt{run(spike\_variance="binomial")} computes the expected busiest core under independent spikes analytically
instead of from mean counts (Appendix~\ref{app:limits}).

\section{Validation details}
\label{app:validation}
Test IDs refer to the pre-registered tests of Tables~\ref{tab:prereg} and~\ref{tab:prereg2} (Appendix~\ref{app:prereg}): V1--V10 validation,
C1--C10 case studies, T1--T20 tool behaviour and limitations.

\begin{figure}[h]
\centering
\includegraphics[width=\linewidth]{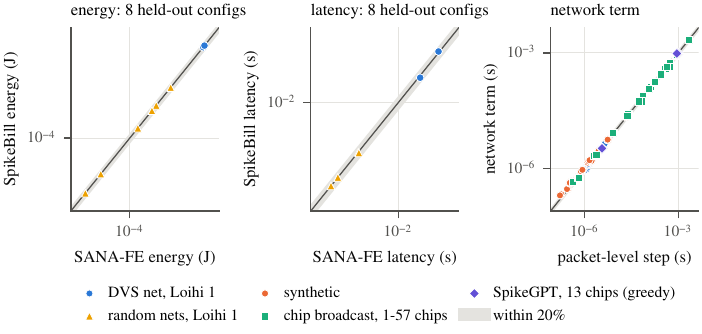}
\caption{Validation. Left, middle: energy and latency per inference of \toolname{} against SANA-FE on the eight
held-out configurations (V2). Right: network time per step against the packet-level simulation for DVS traffic and
synthetic traffic (V4), chip broadcast (V5) and SpikeGPT (greedy mapping, 13 chips). Grey band: $\pm20$\%.}
\label{fig:validation}
\end{figure}

\subsection{Against SANA-FE (Loihi 1)}
\label{app:val-sanafe}
On the eight held-out configurations (V2) the model is within 0.25\% of SANA-FE in energy and 6.2\% in total latency,
and per-step latency is within 20\% on 84\% of the DVS steps. Both simulators used identical chip parameters, mappings
and activity on SANA-FE's public DVS-gesture network and on random networks. We developed the model on three DVS
mappings and nine random networks (V1); V1 failed its criterion before the two changes whose effect the next paragraph measures.

The latency error moves by up to one percentage point between repeated runs, because SANA-FE's detailed timing varies
from run to run; the energy error does not move. Ablations on the held-out set attribute the agreement to two
choices: overlapping the two sides of a core (summing them gives latency errors of $+3$ to $+86$\%) and weighting flows
by per-neuron rates (without it, energy is up to 7.4\% low).

The near-zero energy error of V2 mainly confirms that we count events as SANA-FE does, because the activity came from
SANA-FE's run. V3 closes this loop: our functional simulator produces the activity on five mappings, with the result
given in Section~\ref{sec:validation}.

\subsection{Against a packet-level simulation}
\label{app:val-packet}
\textbf{DVS and synthetic traffic (V4).} The network term over-predicts only at light load. On the DVS network on
Loihi~1 it is within $-16$ to $+4$\% of the packet simulation (median $+2.5$\%). On synthetic traffic on $2\times2$ and
$4\times4$ boards of Loihi-2-class chips, the median error is $+4$\%. The patterns are uniform, hotspot and
bit-complement (each node sends to the node whose address bits are inverted, so traffic crosses the middle of the
board), each at three loads. 17\% of these steps are outside $\pm20$\%, all at the lowest load (4 packets per core),
where the model over-predicts by 16--31\%. At 64 packets per core the error is at most 4\%.

\textbf{SpikeGPT.} On SpikeGPT, mapped greedily on 13 chips, the network term is within $-7.9$ to $+7.7$\% of the
packet simulation on six steps (one mean step, unicast and tree multicast, three seeds). This is about the seed spread
of the packet simulation itself ($\pm$7\% on unicast). The criterion was set after a first look at one seed
(post-hoc).

\textbf{Chip broadcast (V5).} With chip broadcast on 1--57 Loihi-2-class chips, all 22 sampled steps are within 20\%;
the largest error, 17.2\%, occurs at light load, where the unicast step is off by the same amount. The routing mode
matters: on 57 chips a four-layer network takes 921\,\textmu s per step with unicast, 910\,\textmu s with tree
multicast and 118\,\textmu s with chip broadcast in the packet simulation.

\textbf{Finite buffers (V6).} Small router buffers make the model optimistic. The packet simulator checks the
aggregation, not a Loihi~2 network-on-chip, and by default has infinite buffers. With finite link queues and
back-pressure (a router sends only while the next router's queue has room), the share of steps within 20\% is 83\%
with infinite buffers, 78\% with 16-flit queues, 57\% with 4-flit queues and 0\% with 2-flit queues. At 4 flits the
model is optimistic by up to 59\%, mostly for bit-complement traffic, because head-of-line blocking (a blocked packet
at the head of a queue stalls those behind it) is not in the busiest-link bound. Loihi~2's buffer depths are not
public.

\textbf{BookSim2 (V7).} Against BookSim2~\cite{jiang2013booksim}, an unmodified cycle-level simulator, the network
term passes with two virtual channels and fails with the pre-registered one. The configurations are $k\times k$
meshes with $k=8$, 16 and 32, routing first along $x$ then $y$, single-flit packets, and uniform, neighbour, transpose
and bit-complement traffic at 4, 16 and 64 packets per node. From release 0.2.1 the comparison script ships as
\texttt{examples/booksim\_compare.py}; BookSim2, a separate C++ build, is not bundled and is passed with
\texttt{-{}-booksim}. %
With two virtual channels (separate queues that let a blocked
packet be overtaken) and 64-flit buffers, 30 of 36 configurations are within 20\% (median $+0.1$\%); 16-flit buffers
give 81\% and 4-flit buffers 67\%. With one channel, BookSim2 re-allocates the channel for every single-flit packet,
halves its throughput, and 42\% of configurations are within 20\%. We chose two channels after the $k=8$ runs
(post-hoc), and for the same reason dropped a weighted hotspot pattern that a BookSim2 configuration cannot express.

The misses are systematic. Light-load transpose and bit-complement are over-estimated by up to 45\%, because the
busiest link and the longest route belong to different flows; heavy-load bit-complement is under-estimated by up to
35\%, because many links near the maximum load interfere. When all traffic goes to one node, the limit is that node's
input port: BookSim2 sees it (2\x{} the network term), while \toolname{} counts it as the receiving core's work in the
compute term. Our packet simulator is within 20\% of BookSim2 on all 36 configurations, which supports its use as the
reference in V4, V5 and V6.

\subsection{Statistical uncertainty of the pass rates}
\label{app:val-stats}
Table~\ref{tab:val-stats} gives a Wilson 95\% interval for each fraction within $\pm$20\%. SANA-FE's spikes and energy
are deterministic, and its latency varies by up to one point between runs (V2 uses one network generator seed). The
synthetic-traffic and SpikeGPT packet-level checks use three seeds, the chip-broadcast check two, and each BookSim2 value in V7 is the
mean of three seeds. Steps of one run share a network and consecutive inputs, so they are correlated and the
step-level intervals of V2 and V4 are too narrow; the per-configuration counts (all eight held-out configurations
pass V2) are the conservative unit. The intervals of V4 on synthetic traffic (71--91\%) and V7 (68--92\%) and the six SpikeGPT steps do not exclude a
true rate below 75\%, although the observed rates pass.

\begin{table}[h]
\centering
\caption{Fractions within $\pm$20\% of the reference, with Wilson 95\% intervals.}
\label{tab:val-stats}
\small
\begin{tabular}{@{}llrrl@{}}
\toprule
Test & Unit & Within & Rate & 95\% interval \\
\midrule
V2, DVS & steps & 1687 / 2000 & 84.4\% & 82.7--85.9\% \\
V2, random networks & steps & 596 / 600 & 99.3\% & 98.3--99.7\% \\
V4, DVS & steps & 50 / 50 & 100\% & 93--100\% \\
V4, synthetic & steps & 45 / 54 & 83.3\% & 71--91\% \\
V5, chip broadcast & steps & 22 / 22 & 100\% & 85--100\% \\
SpikeGPT packet check & steps & 6 / 6 & 100\% & 61--100\% \\
V7, network term, 64-flit buffers & configurations & 30 / 36 & 83.3\% & 68.1--92.1\% \\
V7, 16-flit buffers & configurations & 29 / 36 & 80.6\% & 65--90\% \\
V7, 4-flit buffers & configurations & 24 / 36 & 66.7\% & 50--80\% \\
V7, one VC (pre-registered) & configurations & 15 / 36 & 41.7\% & 27--58\% \\
V7, our packet simulator & configurations & 36 / 36 & 100\% & 90--100\% \\
\bottomrule
\end{tabular}
\end{table}

\subsection{Direction of the errors}
\label{app:val-signed}
The step-time error has no fixed sign (Table~\ref{tab:signed}). With ample router buffers the network term is
centred, with medians within $\pm$5\%, and runs high at light load, where the busiest link and the longest route
belong to different flows. It runs low when queues are short or many links are near saturation, because the
busiest-link bound ignores that contention. Core time on one chip is within $-6$ to $+4$\% of SANA-FE. This analysis
is descriptive, recomputed from the saved results.

\begin{table}[h]
\centering
\caption{Signed errors, (\toolname{} $-$ reference) / reference. Share low: fraction of cases where \toolname{} is
below the reference.}
\label{tab:signed}
\footnotesize
\setlength{\tabcolsep}{3pt}
\begin{tabular}{@{}llrrrrr@{}}
\toprule
Test & Reference & $n$ & Min & Median & Max & Share low \\
\midrule
V2 total latency & SANA-FE & 8 & $-5.8$\% & $-0.4$\% & $+3.6$\% & 88\% \\
V2 energy & SANA-FE & 8 & $-0.2$\% & $-0.1$\% & 0.0\% & 100\% \\
V4 network, DVS & packet sim. & 50 & $-16.1$\% & $+2.5$\% & $+3.9$\% & 6\% \\
V4 network, synthetic & packet sim. & 54 & $-3.7$\% & $+4.1$\% & $+30.6$\% & 13\% \\
V5 chip broadcast & packet sim. & 22 & $-17.2$\% & $-0.2$\% & $+9.3$\% & 64\% \\
SpikeGPT, 13 chips & packet sim. & 6 & $-7.9$\% & $-5.9$\% & $+7.7$\% & 67\% \\
V6 synthetic, 16-flit queues & packet sim. & 54 & $-35.7$\% & $+3.1$\% & $+26.7$\% & 41\% \\
V6 synthetic, 4-flit queues & packet sim. & 54 & $-58.7$\% & $-18.0$\% & $+8.5$\% & 93\% \\
V7 network, 64-flit buffers & BookSim2 & 36 & $-34.5$\% & $+0.1$\% & $+45.1$\% & 50\% \\
V7 network, 4-flit buffers & BookSim2 & 36 & $-46.1$\% & $-1.3$\% & $+45.1$\% & 53\% \\
\bottomrule
\end{tabular}
\end{table}

\subsection{Loihi 2 calibration}
\label{app:val-loihi2}
The chip file uses $s=0.65$ with a range of 0.33--1.0, which is the spread of our evidence, not a proven bound. Two
of the three PilotNet runs agree at $s=0.63$--0.65, within the $\pm5$\% error of reading the counters off a plot. The
third run, $s=0.33$, comes from a different software revision of the same notebook; it runs at the same frame rate
but reports half the dynamic power. We cannot explain the difference, so the chip file uses it as the low end. The
high end, 1.0, assumes no improvement over Loihi~1.

One factor for all terms is the most the data allow. The three runs share one set of counters, so they give one
equation for four unknown per-term energies. Synaptic events carry 77\% of the energy at Loihi~1 values, so $s$
mainly constrains the synaptic event energy: at $s=0.65$, the scale of the synaptic event energy alone lies between
0.54 and 0.84 for any non-synaptic energies between zero and their Loihi~1 values.

The failed N-DNS transfer is unlikely to come from the neuron cores. The residual, 3.2\,mW, would require 245\,pJ
extra per synaptic operation or 25\,\textmu J per step, and an idle-point measurement on a different Loihi~2
workload~\cite{parpart2023lca} argues against a per-step cost of that size in the neuron cores.

\textbf{A second and a third workload (V9, V10).} A pre-registered fit of the same model to three published
Loihi~2 workloads tests whether the calibration transfers. Dynamic energy per step is modelled as $s$ times the
Loihi~1 per-event energies times the counts, plus a per-core term. The LCA workload is a one-layer sparse-coding
network on 33 cores with an activity sweep at fixed mapping~\cite{parpart2023lca}; we read its active fractions (0,
17\% and 90\%) from the paper's text, not from counters. Fitted alone, LCA gives $s=0.56$ and PilotNet $s=0.64$, a ratio
of 1.15, within the 1.5 criterion; with the low-power PilotNet run the ratio is 1.70. Leaving one workload out fails:
held out, N-DNS is predicted 12\x{} too low, and it would need 13\x{} the LCA scale. N-DNS's published total, 28.7\,mW,
is below the lowest measured Loihi~2 static power (0.15\,W per chip), so its published static/dynamic split is not a
whole-chip split.

\subsection{Speed}
\label{app:val-speed}
A full estimate (mapping, load balancing and evaluation at the value, low and high parameter ends) of the largest
model, Spikingformer with AND-gate attention ($3.35\times10^8$ neurons on 1406 chips), takes 179\,s on eight workers of
a 64-core CPU. Table~\ref{tab:speed} gives the time for a full estimate and per step of a recorded trace.
\begin{table}[h]
\centering
\caption{Time for one full estimate (mapping, balancing, three parameter sets) and per step of a 1000-step trace.}
\label{tab:speed}
\small
\begin{tabular}{@{}lrrrr@{}}
\toprule
Model & Neurons & 1 worker & 8 workers & Trace/step \\
\midrule
SDT V1 8-512 & $1.74\times10^7$ & 130.8\,s & 27.5\,s & 2.2\,ms \\
Spikingformer, AND-gate & $3.35\times10^8$ & 825\,s & 179\,s & 8.8\,ms \\
\bottomrule
\end{tabular}
\end{table}
Compilation alone takes 16\,s for the largest model; most of the time goes to the balancing search (16 targets, two
iterations), which parallelises across targets. On the eight held-out V2 configurations SANA-FE takes 0.21--22.6\,s
and \toolname{} 3--109\,ms (74--3268\x{}, median 483\x{}); on random networks of 8k--131k neurons on one Loihi~1 chip,
3.6--42.5\,s against 0.06--0.34\,s (57--159\x{}). The comparison is not like for like: SANA-FE also simulates the
neurons and so produces the activity, while \toolname{} takes activity as input. The speed-up is for the cost
estimate given activity, which is what design sweeps and estimates inside a training loop repeat.

\section{Spiking-transformer case: details}
\label{app:transformers}

\textbf{Reproducing the reported estimates.} Our graphs reproduce two of the three reported estimates with the
papers' own counting convention: Spikingformer's 7.46\,mJ within 0.05\% and SDT~V1's 4.50\,mJ within 7.7\%. SDT~V2-31M's 32.8\,mJ cannot be
reproduced from its own formula and rate table. The closest value we could obtain from them is 22.4\,mJ, which our
graph matches; the ratios in Table~\ref{tab:transformers} use the reported value, row (a).

\textbf{Alternatives.} The ratios stay above 10\x{} under every alternative we tried. Building attention from AND-gate neurons instead of the co-processor raises them
to 85\x{}, 392\x{} and 52\x{}. Greedy mapping instead of balancing overstates the energy by up to 1.34\x{}. Sampling
all chip parameters jointly gives 26--180\x{} (5th--95th percentiles, assumed log-uniform priors). Under a floor that no real mapping reaches (every weight stored once on 17--24 chips, every
parameter at its favourable end, static power over $T$ steps only), the three models still cost 26\x{}, 14\x{} and
10\x{} their reported estimates.

\textbf{Attention.} How attention's spike--spike products are costed matters only if they are built from neurons.
We costed them in three ways: excluded, on the co-processor, or as AND-gate neurons. The first two agree within 0.1\%,
because the co-processor's energy per operation is small. With AND-gate neurons, Spikingformer needs 1406 chips and
$3.35\times10^8$ neurons; its energy grows 9\x{} and its chip count 8\x{}.

\textbf{Co-processor throughput.} Co-processor throughput matters most for Spikingformer. At nominal parameters the
network sets SDT~V2-31M's step and SDT~V1's cores and network tie, while Spikingformer's co-processor time
(21.5\,\textmu s) sets its step, just above the cores' compute (20.0\,\textmu s) and the network (19.6\,\textmu s). The nominal co-processor is eight shared units at 1\,ns per operation each.
Table~\ref{tab:coproc} compares three configurations, each with the other parameters at their nominal values. With a
single unit, Spikingformer's estimate more than doubles, because its attention products then set the step; one
unit per attention block is slower per block than eight shared units, and the slowest block sets the step. The high
end of Table~\ref{tab:transformers} includes the single unit: with eight units there, Spikingformer's high end is
134\x{} instead of 202\x{}.

\begin{table}[h]
\centering
\caption{Energy change of the chip estimate for three co-processor configurations, relative to the nominal one
(eight shared units at 1\,ns per operation), other parameters nominal.}
\label{tab:coproc}
\small
\begin{tabular}{@{}lrrr@{}}
\toprule
Co-processor & SDT V1 & Spikingformer & SDT V2-31M \\
\midrule
Eight shared units (nominal) & 0 & 0 & 0 \\
One shared unit (the parameter's high end) & 0\% & $+127$\% & $+20$\% \\
One unit per attention block & 0\% & $+8$\% & 0\% \\
\bottomrule
\end{tabular}
\end{table}

\textbf{Memory read per event.} The 45\,nm memory-read reference of row (c$'$) depends on the memory size and on
whether events share reads. With an 8\,KB SRAM instead of 32\,KB, the read accounts for 48\% of row (d) instead of
91\%; if eight events shared one read, it would account for 10\%. Over the parameter range the per-event factor
(c$\to$d) is 11--37\x{}.

\textbf{Mapping.} Balancing lowers the ratios, so the greedy alternative would make the result stronger, not weaker.
For SDT~V1 it shortens the step from 68.1 to 40.0\,\textmu s and lowers the energy from 416 to 353\,mJ, so the ratio
drops from 92\x{} to 78\x{}. For Spikingformer greedy mapping gives 57\x{} instead of 42\x{}, with a 2.3\x{}
longer step. Letting populations share cores would save at most one chip, so the one-population-per-core rule does not
set the chip count. Searching the placement as well with split blocks shortens SDT~V2-31M's step by a further 13\% and changes the other two
by under 0.2\% (Appendix~\ref{app:limits}).

\textbf{Layer variants.} Layer choices that an operation count cannot see change the chip estimate. SDT~V2-31M's RepConv
is trained as a 1\x{}1, depthwise and 1\x{}1 sequence and fused into one dense 3\x{}3 convolution for inference. Kept
unfused, it lowers SDT~V2-31M's estimate from 970 to 821\,mJ (25\x{} the reported estimate), because the fused form stores and reads
more weights.

\textbf{Chip parameters.} One parameter at a time, the synaptic event energy sets the low side of the range and the
time parameters the high side. Moving each parameter between its low and high end with the others nominal, the
synaptic event energy changes the result by $-25$ to $-30$\% and $+27$ to $+33$\%. The synaptic event time raises it
by $+59$ to $+73$\%, the link timing by up to $+65$\% and the chip-to-chip hop time by $+46$ to $+57$\%; slower times
stretch every chip step, and static energy grows with the step. Static power per chip (0.15--0.55\,W) changes the
result by $-10$ to $+42$\%, and every other parameter by at most 20\%, except the co-processor throughput
(Table~\ref{tab:coproc}). 

With the lowest calibration fit ($s=0.33$) as the nominal value, the three models cost 52\x{}, 28\x{} and 19\x{} their reported estimates. Better public numbers
for three Loihi~2 parameters (synaptic event energy, synaptic throughput, link throughput) would narrow these ranges
by most of their width.

\textbf{Measured rates.} Measured on ImageNet with SDT~V1's released weights, the firing rates reproduce the
published ones. The max-pool rates, which the paper does not report and our graph estimated, are lower when measured;
they lower the estimate by 5\%, and chip semantics lowers it by a further 2\%.

\textbf{Small single-chip SNNs.} The gap also appears on one chip. This analysis is descriptive, computed afterwards
from saved results, not pre-registered. For the trained MNIST MLP (784-256-10) and three-layer convolutional SNN of
Section~\ref{sec:semantics}, in chip semantics over 25 steps, the 45\,nm estimate is 0.60 and 1.27\,\textmu J per
inference. The chip estimates are 48.9 and 99.1\,\textmu J on Loihi~1 (81\x{} and 78\x{}) and 55.7 and 164.5\,\textmu J
on Loihi~2 (92\x{} and 130\x{}), each on one chip. Static power is 30--73\% of these estimates, because a small model
leaves its chip mostly idle; without it the ratios are 31--55\x{}. The static share assumes that the model has the
chip to itself, the Loihi~2 energies are the scaled proxies, and counting the first layer as multiply-accumulates
(4.6\,pJ) would make the 45\,nm estimate 2--5\x{} larger.

\textbf{Joint uncertainty.} The Monte Carlo of Table~\ref{tab:transformers} samples all parameters over their ranges
(512 samples, log-uniform priors, the six per-event energies sharing one quantile because they share the factor $s$).
Triangular priors give narrower bands: 85--171\x{}, 50--116\x{} and 31--63\x{}. The medians lie above the nominal
values because the nominal time parameters sit at the fast end of their ranges. The priors and the independence are
assumptions.

\section{Limitations: details}
\label{app:limits}

\begin{table}[h]
\centering
\caption{Chip estimate with framework-semantics activity relative to chip-semantics activity, same weights and input
spikes (Section~\ref{sec:semantics}). The trained conv SNN reaches 94.1\% (framework) and 92.2\% (chip semantics) on
256 MNIST test images.}
\label{tab:semantics}
\small
\begin{tabular}{@{}llrr@{}}
\toprule
Network & Chip & Energy & Latency \\
\midrule
DVS conv net (C6), 5 mappings & Loihi 1 & +74 to +80\% & +76 to +137\% \\
Trained conv SNN, 3 conv + FC (C7) & Loihi 1 & +31\% & +37\% \\
 & Loihi 2 & +13\% & +7\% \\
Trained MLP 784-256-10 (C8) & Loihi 1 & +0.5\% & 0\% \\
 & Loihi 2 & +0.3\% & 0\% \\
\bottomrule
\end{tabular}
\end{table}

\textbf{Static power.} Measured Loihi~2 static power ranges from 0.15 to 0.55\,W per chip across published
setups~\cite{abreu2025neuromorphic,lavadl2023pilotnet,parpart2023lca}; the chip file spans this range, and the nominal
0.23\,W is one board measurement.

\textbf{Network term.} At light load the model over-predicts network time by 16--31\% because it adds path latency to
link serialisation (V4). The model treats the network-on-chip as one flit per packet without virtual channels. The outer
maximum assumes that compute and communication overlap. With 4-flit queues the model is optimistic by up to 59\%
(Appendix~\ref{app:val-packet}).

\textbf{Spike variance.} Drawing per-core spike counts at random (independent spikes) changes the expected step by
at most 0.9\% on the three transformers. One term is less reliable. When balancing spreads a hot population over
cores that each hold a few of its neurons, the busiest of these cores can send several times its mean-count load;
where such a core sets the step, the model understates it.

The option \texttt{spike\_variance="binomial"} computes the
expected busiest core analytically; it matches the sampled value within 0.2\% on the transformers and adds about one
second per evaluated parameter end. Its receive side was changed after the pre-registered version failed its
criterion on a receive-bound case (post-hoc). Correlated (bursty) activity would raise these ratios.

\textbf{Per-image ratio.} The energy ratio to an operation count also varies by input. We compute the chip
estimate and an operation count (synaptic events times 0.9\,pJ, without the first layer's multiply-accumulates) from
the same image's rates, over 5,000 ImageNet images per model. The 5th/50th/95th percentiles of the ratio are
51.7/55.2/64.1 for SDT~V1, 57.8/59.4/60.7 for SDT~V2-55M and 64.8/67.7/71.4 for Spikingformer 8-768; all 15,000
images lie between 49.2 and 76.5 (T19, T20). At mean rates the ratio equals the per-image median for the last two
models. For SDT~V1 it is 50.6, 8\% below its median, because SDT~V1's static energy follows its latency tail.
Mean-rate ratios therefore understate SDT~V1's typical ratio. This count differs from the reported convention, so
these ratios are not those of Table~\ref{tab:transformers}.

\textbf{Comparison with Intel's mapper.} We could not compare core allocations with Intel's: the open-source Lava
framework~\cite{lava} compiles only for CPU, and its Loihi back end is available only to members of Intel's research
community.

\textbf{Mapping and placement.} The placement search matters most for the network-bound SDT~V2-55M. Greedy
and balanced mappings differ by up to 1.3\x{} in energy and 2.3\x{} in latency for the transformers. With each population
on one contiguous block of cores, the placement search (Appendix~\ref{app:model}) changes the three transformers and
SpikeGPT by at most 2.2\% (a shorter step for SDT~V2-31M, 57.89 to 56.60\,\textmu s). It lowers the EDP of SDT~V2-55M
by 36--38\%, so that model's headline numbers are upper estimates, by 1.15\x{}
in energy and 1.41\x{} in latency. Against the best contiguous placement, splitting a population over
non-contiguous blocks lowers the EDP by 14--16\% for SDT~V2-31M (the range is over two runs of the contiguous search), 6.5\% for SDT~V2-55M
 and under 0.2\% for the other models. SDT~V2-31M's numbers in Section~\ref{sec:transformers} use the graph-order
placement of the released reference. Against it, a searched placement with split blocks shortens SDT~V2-31M's
step by 13\% and lowers its energy by 1.8\% (EDP $-14.6$\%).
Each report states the spread of the balanced mappings that lie within 5\% of the best EDP.

\textbf{Chip size.} The chip file can be resized: the same Loihi~2 cores packed into hypothetical chips of 64 to 8192
cores (T17, T18). With static power counted per core, the transformers' energy changes by 1.6--5.9\% between 128 and
8192 cores per chip, because network energy is 1.5--3.5\% of their total. SpikeGPT costs 27\% more on one
8192-core chip, since 75\% of its cores sit idle but powered. Larger chips shorten the step of the network-bound
SDT~V2-31M: its latency falls to 0.74\x{} at 2048 cores per chip. If half of the static power were fixed per chip,
2048-core chips would save 6--16\% of the energy. No chip of these sizes exists, so these are model outputs only.

\textbf{SpikeGPT port.} Our port of SpikeGPT's released forward pass reaches a perplexity of 215 on WikiText-103
(512-token windows, zero-shot) and gives coherent greedy continuations; no published value exists for this setting.
The pre-registered comparison of the loss with the released code was not run, so the port was checked by reading its
arithmetic against the release line by line.

\textbf{SDT~V2-55M under other rates and mappings.} Our graph has the released checkpoint's 77 spiking
populations, with the same sizes, and our port of the checkpoint reaches 79.65\% top-1 on ImageNet. Halving or doubling the assumed firing rates gives 1.37
and 4.19\,J instead of 2.51\,J. A placement search that relieves the busiest links gives 2.18\,J and 7.6\,ms ($-38$\%
EDP), and greedy mapping instead of balancing gives 3.02\,J and 17.8\,ms. With rates measured on ImageNet the full
model costs 2.33\,J and 11.1\,ms on 435 chips. The balancer then packs more compute per core (68 instead of
20\,\textmu s per step). The roofline, which keeps the mapping but spreads traffic evenly over the links, therefore
comes closer to the full model: it is 1.4\x{} low in latency and 13\% low in energy, against 3.2\x{} and 34\% at the
assumed rates, where its latency is 3.39\,ms against the full model's 10.7\,ms.

\section{Intended use, versioning and maintenance}
\label{app:datasheet}

\textbf{Intended use.} \toolname{} estimates the energy and latency of an SNN on a time-stepped, destination-synapse
neuromorphic chip system from a model, a chip file and activity. It is meant for (i) model designers who would
otherwise report operation counts times 45\,nm energies, (ii) comparing models on the same chip file, and (iii)
architects asking which part of a chip system sets the cost (cores, links, barrier, static power, mapping).

\textbf{Out-of-scope use.} The estimates are not measurements and do not include host, I/O or embedded-processor
power (Section~\ref{sec:limits}). They should not be used to claim that an SNN is more or less efficient than an ANN
on a GPU or accelerator, which the tool does not model, or quoted without the parameter range and the options of the
reporting practice (Section~\ref{sec:usage}). Chips whose cores work differently (processor-based, analog or
mixed-signal) need their own core model and validation.

\textbf{Contents of the release.} Besides the library, the release holds four kinds of data. (i) Five chip files, each
parameter with a value, a range and a source, plus a template for new chips and a changelog. (ii) A model zoo of nine
graphs: SDT~V1 8-512, Spikingformer 8-512, SDT~V2-31M, SpikeGPT-216M, SDT~V2-55M, three variants with AND-gate
attention and one with SDT~V2-31M's re-parameterised convolutions (RepConv) left unfused. Each graph is a JSON file listing populations (size, firing rate) and
projections (kind, fan-out, weight bits); its metadata names the source paper, the source of the rates and the graph
conventions. From release 0.2.1, one datasheet (\texttt{docs/zoo.md}) collects these items per model, with the paper's own estimate and the reference estimate. The zoo holds no trained weights and no input data. (iii) Reference estimates for the four base models
(value and range ends, cost split by term, chip version). (iv) The printed numbers of this paper that the
reproduction script checks. The packet-level simulator and the tests are included as code. From release 0.2.1, \texttt{croissant.json} describes the zoo graphs as Croissant metadata, validated with \texttt{mlcroissant}.

\textbf{Versioning.} Every chip file has a version string and a content hash, and each report prints both. A change to
a chip file bumps its version, and the shipped reference estimates are recomputed with each version, so a published
number can be reproduced with the chip file it names. This paper's results were computed with Loihi~2 v2026.11.2-e6; \toolname{}
0.2.1 ships v2026.11.5-e6, which changes only source text and gives identical reference estimates.

\textbf{Interface.} The command line repeats the Python call (\texttt{spikebill model.nir -{}-chip loihi2 -{}-rates rates.json -{}-range}), and
\texttt{spikebill -{}-chip loihi2 -{}-describe} prints every chip parameter with its range, meaning and source.
\texttt{rep.by\_population} marks the populations on the slowest core, so a model designer sees which layer to
change. \texttt{sim.fast(rates)} re-evaluates new rates in milliseconds, for use inside a training loop, and
\texttt{Simulator.spread} (\texttt{-{}-spread rates.npz}) reports the spread of the cost over per-input rates. \texttt{Simulator.from\_torch(net, x)} traces a PyTorch network and measures its
firing rates, mean and per input. A new chip is one YAML file; the release ships Loihi~1, Loihi~2, a 3D-stacked variant and two toy chips. From release 0.2.1, \texttt{spikebill -{}-benchmark -{}-chip new.yaml} scores a new chip file on every zoo model at the reference settings and reports its energy, latency and EDP relative to the shipped Loihi~2 reference; these ratios are to our Loihi~2 proxy, not to silicon.

\textbf{Reporting block.} \texttt{rep.paper\_block()} prints
items (1)--(3) of the reporting practice (Section~\ref{sec:usage}), except the weight precision, as a LaTeX or Markdown
block, and cites the chip file by version and hash. The author adds the weight precision and the semantics and data of
the activity.

\textbf{PyTorch tracer.} \texttt{Simulator.from\_torch} costs products of two spike tensors as AND-gate neurons or on
the co-processor. It lists other non-spiking operations, such as softmax and LayerNorm, with their element counts and
a warning, and does not cost them; a hand-built graph can assign them to the co-processor, as our SpikeGPT graph does.
On Spikingformer the tracer failed its pre-registered test (T1): one-to-one synapses differed by 72\%, and the
balanced cost by 11--13\%, because our hand-built graph lacked the attention edges that the tracer found. After we
added them to the hand-built graph and changed the balancer's search range (both post-hoc), the balanced costs agree
within 2.4\%.

\textbf{Reproduction.} The model zoo holds the three transformers, SpikeGPT-216M, SDT~V2-55M, and variants with
AND-gate attention or with RepConv left unfused. One script, \texttt{examples/reproduce\_paper.py}, recomputes 98 printed numbers of the case
studies (109 with \texttt{-{}-validate}) from the shipped models and checks each against its printed value, listed in
\texttt{examples/expected/paper\_numbers.json}. The shipped
reference evaluates the parameter ends on the mapping balanced at the value, whereas Table~\ref{tab:transformers}
rebalances at each end. The shipped high end for SDT~V1 is therefore 1293\,mJ against 1199\,mJ (266\x{} the reported 4.5\,mJ, Table~\ref{tab:transformers}). With \texttt{-{}-validate} the
script also reruns SANA-FE and the packet-level simulator and checks 11 printed numbers of
Section~\ref{sec:validation}: the four error maxima as bounds, because SANA-FE's latency varies by up to one point
between runs, and the other seven to the printed digits.

\textbf{Validation command.} \texttt{spikebill-validate}, part of the release, reruns six of the tests
against their criteria and reports pass or fail: the held-out SANA-FE test, the trace test, the packet-level checks
on DVS, synthetic and chip-broadcast traffic, and a packet-level check on SpikeGPT whose criterion was set after the
first seed (post-hoc). A reader can rerun these six tests and \texttt{reproduce\_paper.py} from the release. The other
pre-registered tests of Appendix~\ref{app:prereg} come from seeded experiment scripts in our internal research
repository, each run recorded with its commit; they are not part of the release.

\textbf{Sources and licences.} The code is Apache-2.0. Every chip parameter names its source (a publication, SANA-FE's
silicon calibration, a public measurement notebook, a derivation or an assumption) in the chip file and in
Tables~\ref{tab:params-l1} and~\ref{tab:params-l2}. The chip files take only published numbers from
lava-dl~\cite{lavadl2023pilotnet} (BSD-3-Clause) and SANA-FE~\cite{boyle2025sanafe} (GPL-3.0), each cited. They copy
no code or data files from either, and the release's \texttt{NOTICE} file states this for SANA-FE. Loihi is a trademark of Intel
Corporation. This work is not affiliated with or endorsed by Intel, and the Loihi~2 chip file is our approximation
from public data, not Intel's specification.

\textbf{Maintenance.} The release is maintained by the authors' group. New chip files arrive as pull
requests from a template that requires a value, a range and a source for every parameter. A maintainer reviews them,
and a new or changed file bumps its version and adds an entry to the chip changelog. Continuous integration runs the
tests on Python 3.10 and 3.12 (\texttt{CONTRIBUTING.md} in the release). From release 0.2.1, a chip file whose
per-event energies are calibrated on measurements also reports its leave-one-workload-out errors
(\texttt{calibrate.holdout}, the method of Section~\ref{sec:validation}, with a factor-2 check).

\section{Pre-registered tests and their outcomes}
\label{app:prereg}

Tables~\ref{tab:prereg} and~\ref{tab:prereg2} list every pre-registered test on the models of this paper, refuted or not, with the same
weight. One pre-registered check was not run: the
comparison of our SpikeGPT port's loss with the released code (Appendix~\ref{app:limits}). Each test had its
question, refutation criterion and command committed before the run; changes made after seeing data are marked
post-hoc. Tests are numbered by group: V for validation, C for the case studies, T for tool behaviour
and limitations. Results without an ID are descriptive.

\begin{table}[t]
\centering
\caption{Pre-registered tests. Each hypothesis is what we expected before the run; ``not refuted'' means the
criterion did not reject it. Last column: what the outcome means for the tool or its use (+ supports it, $-$ limits
it, $\cdot$ neutral).}
\label{tab:prereg}
\scriptsize
\setlength{\tabcolsep}{3pt}
\begin{tabular}{@{}>{\raggedright\arraybackslash}p{10mm}>{\raggedright\arraybackslash}p{35mm}>{\raggedright\arraybackslash}p{33mm}>{\raggedright\arraybackslash}p{36mm}lc@{}}
\toprule
ID & Hypothesis & Criterion (refuted if) & Result & Verdict & \\
\midrule
\multicolumn{6}{@{}l}{\emph{Validation (Section~\ref{sec:validation}, Appendix~\ref{app:validation})}}\\
V1 & Aggregated model reproduces SANA-FE (Loihi 1), development set & any config outside energy $\pm$10\% or latency $\pm$20\%; per-step latency within 20\% on $<$75\% of steps & energy $-12.0$\% on one random net; latency up to $+67$\%; fraction of steps within $\pm$20\%: 0.71 (DVS), 0.12 (random) & refuted & $-$ \\ %
V2 & Same, held-out configs & same as V1 & energy $\le$0.25\%; latency $-6.2$ to $+3.6$\% over three runs, fraction of steps 0.84 / 0.99 (after a post-hoc link fix; no criterion changed) & not refuted & + \\ %
V3 & Functional simulator in chip semantics replaces SANA-FE's trace & spikes differ, or any mapping outside energy $\pm$10\% or latency $\pm$20\% & 365,277 spikes, equal; energy 0.0\%, latency $-1.9$ to $-7.5$\% & not refuted & + \\ %
V4 & Aggregated network term matches packet-level simulation & $>$25\% of steps outside $\pm$20\%, or $>$50\% in any group & none (DVS), 17\% (synthetic) & not refuted & + \\ %
V5 & Same, chip broadcast & same as V4 & 0 of 22 steps outside & not refuted & + \\ %
V6 & Network term holds with 4-flit router queues & $<$75\% of steps within $\pm$20\% & 57\% (4 flits); 78\% (16 flits) & refuted & $-$ \\ %
V7 & Aggregated network term matches BookSim2 & $<$75\% of configurations within $\pm$20\% & 1 VC (pre-registered): 42\%; 2 VCs (post-hoc): 83\%, median $+0.1$\% & mixed & $\cdot$ \\ %
V8 & Loihi 2 calibration on PilotNet predicts N-DNS power & prediction not within 2\x{} for any calibration & 11--22\x{} under & refuted & $-$ \\ %
V9 & Fitted on any two of PilotNet, LCA and N-DNS, the model predicts the third within 2\x{} & any fold off by $>$2\x{} & N-DNS predicted 12\x{} low; other folds 4\x{} over or degenerate & refuted & $-$ \\ %
V10 & PilotNet and LCA, fitted separately, agree on the per-event scale & scales differ by $>$1.5\x{} & $s=0.64$ vs 0.56 (1.15\x{}); 1.70\x{} with the low-power PilotNet run & not refuted & + \\ %
\midrule
\multicolumn{6}{@{}l}{\emph{Case studies (Sections~\ref{sec:multichip}--\ref{sec:semantics})}}\\
C1 & Full model differs from a first-order estimate on public multi-chip SNNs & both models within 0.8--1.25\x{} energy and 0.5--2\x{} latency & SpikeGPT 1.04\x{}; SDT~V2-55M 2.02\x{} & not refuted & + \\ %
C2 & SDT~V2-31M rates are a fair proxy for SDT~V2-55M & energy off by $>$10\% or latency by $>$20\% with measured rates & $-7.0$\% / $+3.7$\% & not refuted & + \\ %
C3 & Chip-level cost of SDT V1 / Spikingformer exceeds the reported estimate by more than 3\x{} & refuted if at least half of the models are within 3\x{} & 85.6\x{} / 53.0\x{} (pre-registered run: greedy mapping, earlier \toolname{} version; sanity-check convention and a mapping fix changed after first numbers, post-hoc; Table~\ref{tab:transformers} gives 78\x{} / 42\x{} balanced) & not refuted & + \\ %
C4 & Same for SDT V2-31M & refuted if at most 3\x{} & 31.3\x{} (earlier \toolname{} version; 30\x{} in Table~\ref{tab:transformers}) & not refuted & + \\ %
C5 & Reported SDT V1 rates represent ImageNet activity & energy with measured rates off by $>$10\% & $-5.0$\% & not refuted & + \\ %
C6 & Framework-semantics activity changes chip cost & all mappings within energy $\pm$10\% and latency $\pm$20\% & energy $+74$ to $+80$\% & not refuted & + \\ %
C7 & Same on a trained 3-layer conv SNN & same & $+31$\% (Loihi 1), $+13$\% (Loihi 2) & not refuted & + \\ %
C8 & Framework semantics change cost of a trained MLP & energy $\le$10\% and latency $\le$20\% & energy $+0.5$\% (Loihi~1) / $+0.3$\% (Loihi~2), latency 0\% & refuted & $\cdot$ \\ %
C9 & Chip semantics change SDT V1's cost & energy within 10\% and latency within 20\% & $-2.1$\% / $-1.9$\% & refuted & + \\ %
C10 & Graded (real-valued) inputs dominate SpikeGPT's synaptic events & spike-driven events $>$50\% & $3.2\times10^{-5}$ spike-driven & not refuted & + \\ %
\bottomrule
\end{tabular}
\end{table}

\begin{table}[t]
\centering
\caption{Pre-registered tests, continued (same columns as Table~\ref{tab:prereg}).}
\label{tab:prereg2}
\scriptsize
\setlength{\tabcolsep}{3pt}
\begin{tabular}{@{}>{\raggedright\arraybackslash}p{10mm}>{\raggedright\arraybackslash}p{35mm}>{\raggedright\arraybackslash}p{33mm}>{\raggedright\arraybackslash}p{36mm}lc@{}}
\toprule
ID & Hypothesis & Criterion (refuted if) & Result & Verdict & \\
\midrule
\multicolumn{6}{@{}l}{\emph{Tool behaviour and limitations (Sections~\ref{sec:usage}--\ref{sec:limits}, Appendix~\ref{app:limits})}}\\
T1 & PyTorch front end reproduces the hand-built Spikingformer graph & structure, cost or balanced cost off by $>$10\% & neurons and synapses $\le$0.03\% but one-to-one synapses $+72$\%; balanced $+11$\% energy, $+13$\% latency, because the hand-built graph lacked the attention edges; within 2.4\% after adding them and changing the balancer's search range (both post-hoc) & refuted & $-$ \\ %
T2 & Mean spike counts per core set the expected step & expected step $>$1.10\x{} that at mean counts & $\le$1.009\x{} (transformers) & not refuted & + \\ %
T3 & Analytic expected busiest core matches Monte Carlo & any case off by $>$2\% & sender-only version: $-9.8$\%; receive side added (post-hoc): $\le$0.47\% & refuted & $-$ \\ %
T4 & SDT V1 per-image cost is narrow & 5th--95th percentile width $>$20\% of median & 51\% energy, 98\% latency & refuted & $-$ \\ %
T5 & Cost at mean rates equals mean per-image cost & off by $>$5\% & $-10.5$\% energy, $-27.8$\% latency & refuted & $-$ \\ %
T6 & Balancing on 95th-percentile rates shortens SDT V1's mean per-image latency & latency gain $<$10\%, or energy $>$5\% higher, on held-out images & $-24.4$\% latency, $-5.4$\% energy & not refuted & + \\ %
T7 & For SDT V1, a uniform balancer margin $m$ (the balancer sizes cores for spikes $\times(1+m)$) recovers half the 95th-percentile mapping's latency gain (T6) & held-out mean latency $>$2.99\,ms (half-way between 3.41 and 2.58\,ms) for every $m\le$0.3 & best 3.15\,ms (margin 0.2) & refuted & $\cdot$ \\ %
T8 & A 20\% margin raises EDP by at most 5\% at nominal rates & EDP up by $>$5\% on any zoo model & up to $+12$\% (SDT V1) & refuted & $-$ \\ %
T9 & SDT V1's input max-pooling rates vary widely across images (heavy-tailed); this recurs in Spikingformer 8-768 & coefficient of variation across images of the firing rate of any of the four max-pooling populations in the input stage $<$0.5 & CV 0.11--0.25 & refuted & $\cdot$ \\ %
T10 & 95th-percentile mapping cuts Spikingformer 8-768's held-out latency & gain $<$10\% & $+2.7$\% latency & refuted & $\cdot$ \\ %
T11 & Non-contiguous core blocks improve EDP by $<$5\% & any case $\ge$5\% & up to 16\% (SDT V2-31M) & refuted & + \\ %
T12 & Shipped graph-order placement within 5\% EDP of a searched one & any model $\ge$5\% & SDT V2-31M $-14.6$\%; others $\le$0.13\% & refuted & $-$ \\ %
T13 & Searched placement improves EDP by $<$10\% over graph order on every case & any case $\ge$10\% & pre-registered run: at most 10.0\% on this paper's models (SDT V2-55M greedy at the threshold), refuted only on a model outside this paper; a calibrated search added later (post-hoc) lowers SDT V2-55M's EDP by 38\% & not refuted & $\cdot$ \\ %
T14 & The best of five Kernighan--Lin orders is $<$2\% better than the placement search & any case $\ge$2\% better & worse than the search on every model in this paper (by 0.01--62\%); refuted only on a model outside this paper & refuted & $\cdot$ \\ %
T15 & Compiled evaluation is at least 10\x{} faster per step than the direct per-step code on SDT V1, with equal results & speed-up $<$10\x{} or any step differs by $>$10$^{-9}$ & 1,099\x{}; identical & not refuted & + \\ %
T16 & Difference-array link loads are at least 10\x{} faster than hop walking for 10$^6$ flows on the 1406-chip mesh, with equal loads & below 10\x{} or any load differs & 31\x{}; equal to $4\times10^{-15}$ & not refuted & + \\ %
T17 & Packing the same Loihi~2 cores into chips of 128--8192 cores changes energy by $<$15\% on every zoo model (static power per core) & any model $\ge$15\% & 1.6--5.9\% on the transformers; SpikeGPT $+27$\% on one 8192-core chip that is 75\% idle & refuted & $-$ \\ %
T18 & SDT~V2-31M latency falls by $\ge$20\% at 2048 cores per chip & fall $<$20\% & $\times$0.74 (5.73 to 4.23\,ms) & not refuted & + \\ %
T19 & The ratio of energy to operation count is stable across images & 5th--95th percentile width $>$20\% of median & 22.5\% SDT~V1; 4.9\% SDT~V2-55M; 9.7\% Spikingformer 8-768 & refuted & $-$ \\ %
T20 & The ratio at mean rates lies within the per-image 5th--95th percentiles & outside for any model & SDT~V1 50.6 below 51.7; the others at their medians & refuted & $-$ \\ %
\bottomrule
\end{tabular}
\end{table}

\textbf{Post-hoc changes that affect reported numbers.} The V2 latencies are after a fix of the busiest-link term
(the per-hop latency had been used as the time per message), which changed no criterion; the receive side of
\texttt{spike\_variance="binomial"} was changed after its pre-registered version failed (Appendix~\ref{app:limits}).

\section{A multi-chip Loihi 2 measurement}
\label{app:abreu}
A published 24-chip Loihi~2 language model~\cite{abreu2025neuromorphic} spends 6.9\x{} the energy per token of its
own single-chip estimate, which shows that multi-chip cost is not the sum of single-chip costs. The paper reports both
a single-chip extrapolation and a 24-chip measurement. In generation, the measured 24-chip energy is 405\,mJ per token against a single-chip
estimate of 59\,mJ (6.9\x{}, our ratio); throughput is 1.7\x{} lower (as the paper states) and power about 4\x{}
higher (derived). Time per step rises by about 20\% beyond four chips.

\toolname{} does not reproduce the power part of this gap. The single-chip estimate already counts the idle
power of 24 chips and totals 4.2\,W, so the measured run (16.8\,W) spends about 12.6\,W more, about 0.5\,W per chip
(our arithmetic). In generation a token passes the 24 blocks in sequence, so the neuron-core work per token is the same
in both cases. The extra power belongs to system activity that models of the neuron cores, ours included, do not represent:
chip-to-chip I/O, the embedded processors and the host interface. We attribute the N-DNS under-prediction
(Appendix~\ref{app:val-loihi2}) to this omitted term, without a test. The multi-chip barrier and link terms of \toolname{} could account for
part of the throughput loss, but the publication gives neither the per-step counts needed to compute it nor the
mapping or per-component power. We therefore use it as motivation and as a limitation, not as a validation target.

\section{The 17 surveyed papers}
\label{app:survey}
We read the energy method of 17 spiking-transformer and spiking-language-model papers from 2023--2026 in full text
(Table~\ref{tab:survey}); the list is a search, not a census. E: 45\,nm operation counts with $E_\mathrm{MAC}=4.6$\,pJ,
$E_\mathrm{AC}=0.9$\,pJ stated; V: a close variant of that convention; O: another method; U: method not confirmed.
None of the 17 reports energy measured on neuromorphic hardware; one (SpikedAttention) adds a memory-access term.

\begin{table}[h]
\centering
\caption{Energy methods of 17 spiking-transformer / spiking-LLM papers.}
\label{tab:survey}
\scriptsize
\setlength{\tabcolsep}{3pt}
\resizebox{\linewidth}{!}{\begin{tabular}{@{}llll@{}}
\toprule
Paper & Venue & arXiv & Method \\
\midrule
Spikformer, Zhou et al. & ICLR 2023 & 2209.15425 & E \\
Spike-driven Transformer, Yao et al. & NeurIPS 2023 & 2307.01694 & E \\
SDT V2, Yao et al. & ICLR 2024 & 2404.03663 & E \\
Spikingformer, C. Zhou et al. & AAAI 2026 & 2304.11954 & E \\
SpikeGPT, Zhu et al. & TMLR 2024 & 2302.13939 & V \\
SpikeLM, Xing et al. & ICML 2024 & 2406.03287 & E \\
SpikingBERT, Bal \& Sengupta & AAAI 2024 & 2308.10873 & E \\
SpikingResformer, Shi et al. & CVPR 2024 & 2403.14302 & V \\
QKFormer, Zhou et al. & NeurIPS 2024 & 2403.16552 & E \\
E-SpikeFormer, Yao et al. & IEEE TPAMI 2025 & 2411.16061 & U \\
SpikeYOLO, Luo et al. & ECCV 2024 & 2407.20708 & E \\
STAtten, Lee et al. & CVPR 2025 & 2409.19764 & E \\
Spike-driven LLM, Xu et al. & arXiv & 2604.16475 & E \\
SpikeZIP-TF, You et al. & ICML 2024 & 2406.03470 & O (AC only) \\
SpikedAttention, Hwang et al. & NeurIPS 2024 & -- & O (AC + memory) \\
SpikeLLM, Xing et al. & ICLR 2025 & 2407.04752 & O (bit-ops) \\
Spikformer V2, Zhou et al. & arXiv & 2401.02020 & V \\
\bottomrule
\end{tabular}}
\end{table}

\section{Chip parameters}
\label{app:params}
Tables~\ref{tab:params-l1} and~\ref{tab:params-l2} list every parameter of Eqs.~(1)--(2) for the two chips, with
the chip-file names that Table~\ref{tab:symbols} maps to the symbols of the equations. The values are those of the
chip files in \toolname{} 0.2.1. Our results were computed with Loihi~2 v2026.11.2-e6; the later versions change
only source text. The suffix e6 marks Loihi~2 energies scaled by the fitted factor of Section~\ref{sec:validation}.

Four notes help in reading the tables. $N_{core}$ is the neuron capacity of a core, not $N$ of Eq.~(1). Loihi~1 has
one synaptic event energy per projection type (dense, convolutional, sparse). ``Range'' is the low--high interval
that the range, sensitivity and Monte Carlo analyses use. The source tags are: [P] published, [C] silicon-calibrated by
SANA-FE, [D] derived from published totals, [S] scaled by the fitted factor (Appendix~\ref{app:val-loihi2}), [L1]
Loihi~1 value used as a proxy, [A] assumption.

The co-processor of Appendix~\ref{app:model} is in neither table, because neither chip has one. Its parameters are
assumptions [A]: $e_\mathrm{op}=0.03$\,pJ per operation (range 0.01--0.9\,pJ) and $t_\mathrm{op}=0.125$\,ns per
operation, i.e.\ eight units at 1\,ns each (high end: one unit, 1\,ns).

\begin{table}[h]
\centering
\caption{Symbols of Eqs.~(1)--(2) and the chip-file names in Tables~\ref{tab:params-l1} and~\ref{tab:params-l2}.}
\label{tab:symbols}
\small
\begin{tabular}{@{}ll@{\hspace{8mm}}ll@{}}
\toprule
Eq. & Chip file & Eq. & Chip file \\
\midrule
$e_\mathrm{syn}$ & $e_{synop}$ & $t_\mathrm{syn}$ & $t_{synop}$ \\
$e_\mathrm{upd}$ & $e_{neuron\_update}$ & $t_\mathrm{upd}$ & $t_{neuron\_update}$ \\
$e_\mathrm{act}$ & $e_{neuron\_active}$ & $t_\mathrm{act}$ & $t_{neuron\_active}$ \\
$e_\mathrm{spk}$ & $e_{spike\_gen}$ & $t_\mathrm{spk}$ & $t_{spike\_gen}$ \\
$e_\mathrm{out}$ & $e_{spike\_out}$ & $t_\mathrm{out}$, $t_\mathrm{in}$ & $t_{spike\_out}$, $t_{msg\_in}$ \\
$e_\mathrm{hop}$ & $e_{hop}$ & $t_\mathrm{link}$ & $t_{link}$ \\
$e_\mathrm{chop}$ & $e_{chip\_hop}$ & $t_\mathrm{clink}$ & $t_{chip\_hop}$ \\
$P_\mathrm{chip}$, $P_\mathrm{core}$ & $p_{static\_chip}$, $p_{static\_core}$ & $t_\mathrm{bar}$ & $t_{barrier}$; $t_{barrier}^{multi}$ across chips \\
 & & $t_\mathrm{path}$ & hops of the longest route $\times$ $t_{hop}$ \\
\bottomrule
\end{tabular}
\end{table}

\begin{table}[t]
\centering
\caption{Loihi~1 chip file v2026.11.3 (8$\times$4 tiles, 4 cores per tile).}
\label{tab:params-l1}
\scriptsize
\setlength{\tabcolsep}{3pt}
\resizebox{\linewidth}{!}{\begin{tabular}{@{}lllp{38mm}p{72mm}@{}}
\toprule
Symbol & Value & Range & Meaning & Source \\
\midrule
$N_{core}$ & 1.02e+03 & -- & neurons per core & [P] Davies18 constraint 1 \\
$M_{syn}$ & 1.05e+06 & -- & synapse memory per core (bits) & [P] Davies18 constraint 2 (128 KB; before compression) \\
$A_{in}$ & 4.1e+03 & -- & input axons per core & [P] Davies18 constraint 4 \\
$A_{out}$ & 4.1e+03 & -- & output axons per core & [P] Davies18 constraint 3 \\
$e_{synop}$ & 3.55e-11 J & 2.36e-11--3.55e-11 & energy per synaptic event & [C] SANA-FE dense synapse; [P] Davies18 T2 23.6 pJ (min) \\
$e_{synop}$ & 2.4e-11 J & -- & energy per synaptic event & [C] SANA-FE conv synapse \\
$e_{synop}$ & 3.36e-11 J & -- & energy per synaptic event & [C] SANA-FE sparse synapse \\
$t_{synop}$ & 3.1e-09 s & -- & time per synaptic event & [C] SANA-FE conv synapse \\
$t_{synop}$ & 4.7e-09 s & -- & time per synaptic event & [C] SANA-FE sparse synapse \\
$e_{neuron\_update}$ & 5.12e-11 J & 5.12e-11--5.2e-11 & energy per neuron per step (all mapped neurons) & [C] SANA-FE soma access, every neuron; [P] Davies18 T2 52 inactive \\
$e_{neuron\_active}$ & 2.16e-11 J & 2.16e-11--2.9e-11 & extra energy per neuron with input or state & [C] SANA-FE soma update; [P] Davies18 81 active - 52 inactive \\
$e_{spike\_gen}$ & 6.93e-11 J & 0--6.93e-11 & energy per spike generated & [C] SANA-FE soma spike out; not separate in Davies18 \\
$e_{spike\_out}$ & 1.11e-10 J & 1.7e-12--1.11e-10 & energy per message sent & [C] SANA-FE axon out per message; [P] Davies18 T2 within-tile spike 1.7 pJ \\
$e_{hop}$ & 3.6e-12 J & 3e-12--4.2e-12 & energy per tile-to-tile hop & [C] SANA-FE 3.0 E-W / 4.2 N-S, mean; [P] Davies18 3.0/4.0 \\
$e_{chip\_hop}$ & 3.6e-11 J & 3.6e-12--3.6e-10 & energy per chip-to-chip hop & [A] not public; 10x on-chip hop, range 1x-100x \\
$t_{synop}$ & 3.8e-09 s & 3.5e-09--4.7e-09 & time per synaptic event & [C] SANA-FE dense 3.8 / sparse 4.7 ns; [P] Davies18 3.5 ns \\
$t_{neuron\_update}$ & 6e-09 s & 5.3e-09--6e-09 & time per neuron per step & [C] SANA-FE soma access 6.0 ns; [P] Davies18 5.3 inactive \\
$t_{neuron\_active}$ & 3.7e-09 s & 3.1e-09--3.7e-09 & extra time per active neuron & [C] SANA-FE soma update; [P] Davies18 8.4 - 5.3 \\
$t_{spike\_gen}$ & 3e-08 s & -- & time per spike generated & [C] SANA-FE soma spike out \\
$t_{spike\_out}$ & 5.1e-09 s & 2.1e-09--5.1e-09 & time per message sent (core side) & [C] SANA-FE axon out per message; [P] Davies18 within-tile 2.1 ns \\
$t_{msg\_in}$ & 1.6e-08 s & -- & time per message received (core side) & [C] SANA-FE axon in per message \\
$t_{hop}$ & 5.3e-09 s & 4.1e-09--6.5e-09 & latency per tile hop & [P]/[C] 4.1 E-W / 6.5 N-S, mean \\
$t_{link}$ & 2.9e-10 s & 2.9e-10--5.3e-09 & time per message on a tile link (1 / bandwidth) & [P] Davies18 T2: 3.44 Gspike/s cross-sectional bandwidth per tile $\to$ 1/3.44e9; high = hop latency \\
$t_{chip\_hop}$ & 5.4e-09 s & 5.4e-09--2.16e-08 & time per message on a chip-to-chip link & [D] 1/185 Mtransfers/s (Loihi 2 brief fn 3) per transfer; high = 4 transfers/packet [A] \\
$t_{barrier}$ & 1.8e-06 s & 4.65e-07--1.8e-06 & barrier (one chip; barrier\_table overrides) & [C] SANA-FE latency\_sync 29 cores 1.8 us; [P] Davies18 113-465 ns (1-32 tiles) \\
$t_{barrier}^{multi}$ & 1.43e-05 s & 1.8e-06--1.62e-05 & barrier across chips & [P] Frady et al. 2020 Sec 5.1: 14.3 us (4x64 chips), 16.2 us (768 chips); low = single chip \\
$p_{static\_core}$ & 0 W & -- & static power per used core & static booked per chip; NxSDK pro-rates by used cores (alternative accounting) \\
$p_{static\_chip}$ & 0.0695 W & 0--0.0695 & static power per chip & [D] Frady 2020 Table 1: 53.4 W / 768 chips; low = dynamic-only accounting (SANA-FE) \\
\midrule
barrier table & \multicolumn{4}{l}{1 tile: 0.6 us, 2 tiles: 1 us, 4 tiles: 1.4 us, 29 tiles: 1.8 us} \\
\bottomrule
\end{tabular}}
\end{table}

\begin{table}[t]
\centering
\caption{Loihi~2 chip file v2026.11.5-e6 (8$\times$4 tiles, 4 cores per tile). The per-event energies are Loihi~1 values
scaled by 0.65 (range 0.33--1.0).}
\label{tab:params-l2}
\scriptsize
\setlength{\tabcolsep}{3pt}
\resizebox{\linewidth}{!}{\begin{tabular}{@{}lllp{38mm}p{72mm}@{}}
\toprule
Symbol & Value & Range & Meaning & Source \\
\midrule
$N_{core}$ & 8.19e+03 & -- & neurons per core & [P] brief Fig. 1 (up to 8192; depends on state size) \\
$M_{syn}$ & 1.05e+06 & -- & synapse memory per core (bits) & [P] brief Fig. 1: up to 128 KB synaptic memory per core \\
$A_{in}$ & 4.1e+03 & -- & input axons per core & [A] Loihi 1 value; Loihi 2 not public \\
$A_{out}$ & 4.1e+03 & -- & output axons per core & [A] Loihi 1 value; Loihi 2 not public \\
$e_{synop}$ & 2.31e-11 J & 1.17e-11--3.55e-11 & energy per synaptic event & [S] 0.65 x [L1] (range 0.33-1.0); [L1] SANA-FE dense synapse \\
$e_{neuron\_update}$ & 3.33e-11 J & 1.69e-11--5.12e-11 & energy per neuron per step (all mapped neurons) & [S] 0.65 x [L1] (range 0.33-1.0); [L1] SANA-FE soma access \\
$e_{neuron\_active}$ & 1.4e-11 J & 7.13e-12--2.16e-11 & extra energy per neuron with input or state & [S] 0.65 x [L1] (range 0.33-1.0); [L1] SANA-FE soma update \\
$e_{spike\_gen}$ & 4.5e-11 J & 2.29e-11--6.93e-11 & energy per spike generated & [S] 0.65 x [L1] (range 0.33-1.0); [L1] SANA-FE soma spike out \\
$e_{spike\_out}$ & 7.22e-11 J & 3.66e-11--1.11e-10 & energy per message sent & [S] 0.65 x [L1] (range 0.33-1.0); [L1] SANA-FE axon out per message \\
$e_{hop}$ & 2.34e-12 J & 1.19e-12--3.6e-12 & energy per tile-to-tile hop & [S] 0.65 x [L1] (range 0.33-1.0); [L1] SANA-FE hop mean \\
$e_{chip\_hop}$ & 3.6e-11 J & 3.6e-12--3.6e-10 & energy per chip-to-chip hop & [A] not public; 10x the Loihi 1 on-chip hop (3.6 pJ), range 1x-100x \\
$t_{synop}$ & 7.6e-10 s & 7.6e-10--3.8e-09 & time per synaptic event & [P] brief: 5x Loihi 1 synaptic ops; [L1] 3.8 ns / 5 \\
$t_{neuron\_update}$ & 3e-09 s & 3e-09--6e-09 & time per neuron per step & [P] brief: 2x neuron update; [L1] 6.0 ns / 2 \\
$t_{neuron\_active}$ & 1.85e-09 s & 1.85e-09--3.7e-09 & extra time per active neuron & [P] 2x; [L1] 3.7 ns / 2 \\
$t_{spike\_gen}$ & 3e-09 s & 3e-09--3e-08 & time per spike generated & [P] brief: 10x spike generation; [L1] 30 ns / 10 \\
$t_{spike\_out}$ & 5.1e-09 s & -- & time per message sent (core side) & [L1] axon out (no Loihi 2 figure) \\
$t_{msg\_in}$ & 1.6e-08 s & 3.2e-09--1.6e-08 & time per message received (core side) & [L1] axon in; low assumes 5x like synops [A] \\
$t_{hop}$ & 5.3e-09 s & 4.1e-09--6.5e-09 & latency per tile hop & [L1] hop latency \\
$t_{link}$ & 2.9e-10 s & 2.9e-10--5.3e-09 & time per message on a tile link (1 / bandwidth) & [L1] Loihi 1 tile bandwidth 3.44 Gspike/s; high = hop latency \\
$t_{chip\_hop}$ & 1.25e-09 s & 1.25e-09--5e-09 & time per message on a chip-to-chip link & [P] brief fn 3: 800 Mtransfers/s (simulated); high = 4 transfers/packet [A] \\
$t_{barrier}$ & 1.8e-06 s & 2e-07--1.8e-06 & barrier (one chip; barrier\_table overrides) & [P] brief: min timestep \textless{} 200 ns (low); [L1] 1.8 us (value) \\
$t_{barrier}^{multi}$ & 1.43e-05 s & 1.8e-06--1.62e-05 & barrier across chips & [L1] Frady 2020 (Loihi 1 multichip) as proxy \\
$p_{static\_core}$ & 0 W & -- & static power per used core & static booked per chip \\
$p_{static\_chip}$ & 0.23 W & 0.15--0.55 & static power per chip & [P] lava-dl PilotNet LIF notebook 0.232 W (value); low Abreu 2025 \textasciitilde{}0.15 W; high Parpart 2023 0.54-0.55 W \\
\bottomrule
\end{tabular}
}
\end{table}

\ifnum\anonbuild=1
\input{sections/checklist}
\fi

\end{document}